\pdfoutput=1

\PassOptionsToPackage{colorlinks,linkcolor=ProfessionalOrange,citecolor=ProfessionalOrange,urlcolor=ProfessionalOrange}{hyperref}

\documentclass[11pt]{article}

\usepackage[preprint]{acl}

\usepackage{times}
\usepackage{latexsym}

\usepackage[T1]{fontenc}

\usepackage[utf8]{inputenc}

\usepackage{microtype}

\usepackage{inconsolata}

\usepackage{graphicx}
\usepackage{tabularx}
\usepackage{multirow}

\usepackage{wrapfig}

\usepackage{booktabs}   
\usepackage{longtable}
\usepackage{tabularx}
\usepackage{amsfonts} 

\usepackage{dialogue}

\usepackage{xcolor}
\definecolor{ProfessionalOrange}{HTML}{722F37}

\usepackage[most]{tcolorbox}
\usepackage{enumitem}
\usepackage{verbatim}
\usepackage{fancyvrb}
\usepackage{fvextra}
\RecustomVerbatimEnvironment{Verbatim}{Verbatim}{
  fontsize=\small,
  breaklines=true,       
  breakanywhere=true,    
  xleftmargin=1em,       
  xrightmargin=1em,      
  samepage=false,         
  breaksymbol={},    
  breakindent=0pt,   
}

\usepackage{tabularx}

\newcounter{dialogueturn}
\newenvironment{persuasiondialogue}{%
  \setcounter{dialogueturn}{1}%
  \begin{dialogue}%
}{%
  \end{dialogue}%
}

\newcommand{\persuader}[1]{%
  \speak{\textcolor{red}{Persuader} (\textbf{\textcolor{red}{\arabic{dialogueturn}}})} #1%
  \stepcounter{dialogueturn}%
}

\newcommand{\persuadee}[1]{%
  \speak{\textcolor{blue}{Persuadee} (\textbf{\textcolor{blue}{\arabic{dialogueturn}}})} #1%
  \stepcounter{dialogueturn}%
}

\title{How Do LLMs Persuade? Linear Probes Can Uncover Persuasion Dynamics in Multi-Turn Conversations}

\author{%
  Brandon Jaipersaud$^{1}$,
  David Krueger$^{1,2}$,
  Ekdeep Singh Lubana$^{3}$ \\[1ex]
  ${}^{1}$Mila\\
  ${}^{2}$University of Montreal\\
  ${}^{3}$CBS-NTT Program in Physics of Intelligence, Harvard University\\[1ex]
  \texttt{brandonjaip@gmail.com}
}

\usepackage{minitoc}
\usepackage[toc,page,header]{appendix}
\usepackage[color=pink!20,textsize=scriptsize]{todonotes}

\begin{document}

    \doparttoc 
    \faketableofcontents 

\maketitle

\begin{abstract}
Large Language Models (LLMs) have started to demonstrate the ability to persuade humans, yet our understanding of \textit{how} this dynamic transpires is limited. Recent work has used linear probes, lightweight tools for analyzing model representations, to study various LLM skills such as the ability to model user sentiment and political perspective. Motivated by this, we apply probes to study persuasion dynamics in natural, multi-turn conversations. We leverage insights from cognitive science to train probes on distinct aspects of persuasion: persuasion success, persuadee personality, and persuasion strategy. Despite their simplicity, we show that they capture various aspects of persuasion at both the sample and dataset levels. For instance, probes can identify the point in a conversation where the persuadee was persuaded or where persuasive success generally occurs across the entire dataset. We also show that in addition to being faster than expensive prompting-based approaches, probes can do just as well and even outperform prompting in some settings, such as when uncovering persuasion strategy. This \textit{suggests} probes as a plausible avenue for studying other complex behaviours such as deception and manipulation, especially in \textit{multi-turn settings} and large-scale dataset analysis where prompting-based methods would be computationally inefficient.
\end{abstract}


\section{Introduction}
Large Language Models (LLMs) have started to exhibit the ability to influence users' beliefs and opinions, with an efficacy argued to be comparable to human communicators~\cite{Salvi2024OnTC, luciano2024hypersuasion, carrascofarre2024largelanguagemodelspersuasive}. This phenomenon, termed \textit{LLM-based persuasion}, is a dual-use capability that can have both concerning uses, e.g., political targeting and spreading misinformation~\cite{doi:10.1073/pnas.2403116121, 10.1002/aaai.12188}, as well as beneficial applications, e.g., education and therapy~\cite{garcia2025review, stade2024large, obradovich2024opportunities}. 

Despite the growing evidence that LLMs have persuasive influence on humans, we lack a foundational understanding of \textit{how} this dynamic transpires in a conversation. Specifically, as a rather abstract, high-level behavior, persuasion has generally been studied as an inherently human capacity with a strong grounding in cognitive science literature~\cite{EAGLY1984267}. Dual-process frameworks, for instance, such as the heuristic-systematic model, assert that persuasion is a phenomenon that can be primarily attributed to the recipient of a message \cite{chaiken1980heuristic, greenwald1968cognitive}. Alternative theories emphasize the sender's role \cite{Briñol01022009, d090b47e-e06a-3e94-a275-bb65a2f40518, TORMALA2006684} or message content characteristics \cite{shen2013effects, hoeken2022reconstructability}. These theories help ground behavioral studies, e.g., \citet{OYIBO2019174, ALKIS2015147}, which aim to quantify the interaction effects of factors such as personality and strategy on persuasive success.

\begin{figure}[h]
    \centering
    \includegraphics[width=0.5\textwidth]{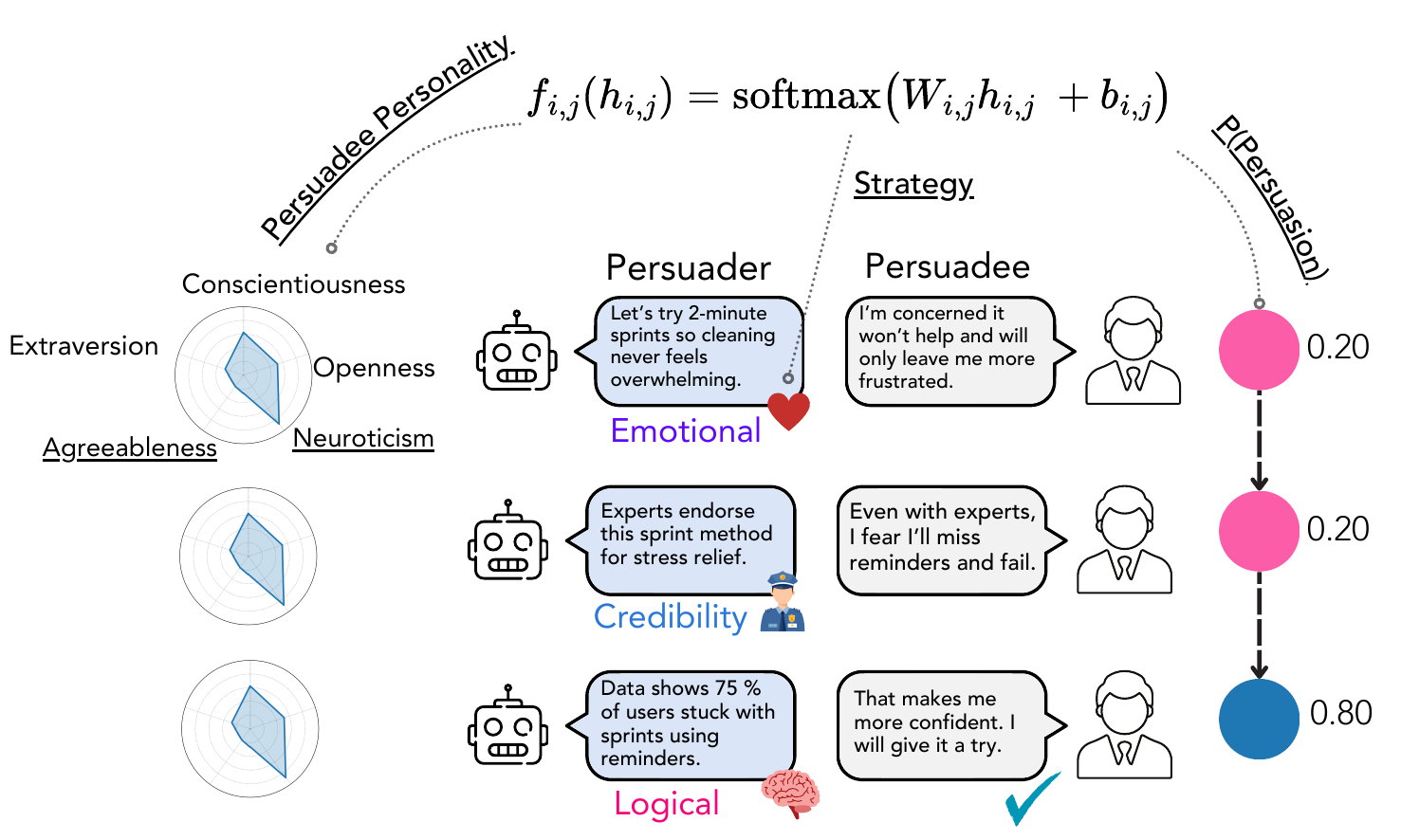}
\caption{
\textbf{Our approach to study how LLMs persuade in natural, multi-turn conversations.} Following insights from cognitive science demonstrating interaction effects of personality and strategy on persuasion, we develop linear probes for these three behavioral dimensions. Our probes, $f_{i,j}$ are logistic regression classifiers which we apply to model activations $h_{i,j}$ extracted from the residual stream output of the $i^{th}$ layer and $j^{th}$ token (see Section~\ref{sec:exp-setup} for detailed notation).} 
    \label{sys-diagram}
\end{figure}

Leveraging these insights from cognitive science literature, which show personality and persuasion strategy as key causal variables that influence persuasive outcomes: \textit{our goal in this paper is to better understand how LLMs can persuade humans in semi-naturalistic, multi-turn settings.} 
As multi-turn dialogue can involve a large volume of tokens, prompting-based analysis becomes infeasible for this goal, as it requires an expensive forward pass for each token-level query. 
To circumvent this challenge, we thus use \textit{linear probes}, a classic tool in NLP research that is often applied to provide token-level insights across various abstract phenomena in LLMs such as sentiment, space, time, and political perspective \cite{tigges2024language, kim2025linear, gurnee2024language}. By using probes, we can efficiently analyse multi-turn dialogue at finer granularities such as at the token and turn levels. To apply probes for studying persuasion dynamics, \textit{we train three specialized probes, each targeting distinct aspects of persuasion:} overall persuasion outcome, personality of the persuadee, and rhetorical strategies (Figure~\ref{sys-diagram}). This approach allows us to efficiently analyse persuasion dynamics, compared to prompting-based methods, an advantage which we quantify in Figure~\ref{fig:timing}. 
Overall, we make the following contributions.
\begin{itemize}[leftmargin=12pt]
    \item \textbf{A framework for analyzing persuasion dynamics in LLM-driven conversations using linear probes.} We design lightweight, efficient probes that capture key aspects of persuasion, enabling fine-grained, turn-level analysis at scale. We find these probes not only match or exceed the performance of prompting-based methods but also provide significant computational efficiency, making them a practical tool for large-scale persuasion analysis.
    
    \item \textbf{Probing persuasion outcomes, rhetorical strategies, and personality traits.} We demonstrate that linear probes trained on LLM activations can accurately identify where persuasion success or failure occurs, detect rhetorical strategies employed by the persuader, and estimate persuadee personality across a conversation.
    
    \item \textbf{Empirical insights into persuasion trajectories across synthetic and human datasets.} We show that persuasive cues concentrate around the middle turns of human dialogue but shift to the final one or two turns in LLM-generated dialogue, revealing a systematic divergence in when persuasion unfolds across natural versus synthetic data.
    
    \item \textbf{Correlations between strategy and personality.} By correlating probe outputs, we reveal that traits like extraversion modulate the effectiveness of different rhetorical strategies (e.g., credibility or emotional appeals), offering a nuanced picture of how LLMs might adapt persuasion tactics.
\end{itemize}

\section{Experimental Setup}
\label{sec:exp-setup}
We are interested in detecting persuasion behaviors in LLMs using linear probes. Each probe performs multi-class logistic regression trained using empirical risk minimization on frozen LLM activations. Concretely, let: 

\begin{equation}
    \begin{aligned}
    \mathcal D &= \bigl\{( h_{i,j}^{(n)},\,y^{(n)})\bigr\}_{n=1}^N,\\
    h_{i,j}^{(n)} &\in \mathbb R^d,\quad
    y^{(n)}\in\{1,\dots,C\}.
    \end{aligned}
\end{equation}

be our training set of $d$-dimensional activations and integer labels $y^{(n)} \in \{1, \dots, C\}$. $i$ denotes the residual stream layer (i.e. the output of the transformer block) while $j$ represents the token index from which we extract activations. Our linear probe $f_{i,j} : \mathbb{R}^d \rightarrow\ \mathbb{R}^C$ computes: 

\begin{equation}
f_{i,j}(h_{i,j})
= \mathrm{softmax}\bigl(W_{i,j }h_{i,j}\ + b_{i,j}\bigr) \in \Delta^{C}
\end{equation}
\begin{equation*}
W_{i,j} \in \mathbb{R}^{C\times d}, 
\quad
b_{i,j} \in \mathbb{R}^C.
\end{equation*}
where $W_{i,j}$, $b_{i,j}$ are our trainable weights  and biases pertaining to the $i^{th}$ layer and $j^{th}$ token, while $\Delta^{C}$ is the $C$-probability simplex. Let $\theta_{i,j} = \{\,W_{i,j},\;b_{i,j}\}$. We optimize $\theta_{i,j}$ using a cross-entropy loss objective (\ref{cross-entropy}, sample-wise) and gradient descent (\ref{gd}): 
\begin{equation}
\label{cross-entropy}
\begin{split}
\ell\bigl(f_{i,j}(h_{i,j}),y\bigr)
&= -\sum_{k=1}^C \mathbf{1}\{y=k\}\,
   \log\bigl[f_{i,j}(h_{i,j})\bigr]_k \\[0.5ex]
&= -\log\bigl[f_{i,j}(h_{i,j})\bigr]_y.
\end{split}
\end{equation}
\begin{equation}
\label{gd}
    \theta_{i,j}^{p+1} \;\leftarrow\;
\theta_{i,j}^{p} - \eta \,\nabla_{\theta_{i,j}^{p}}\!\biggl[\tfrac1N\sum_{n=1}^N
\ell\!\bigl(f_{i,j}( h_{i,j}^{(n)}),y^{(n)}\bigr)\biggr]
\end{equation}

\begin{figure}  
    \includegraphics[width=1.00\columnwidth]{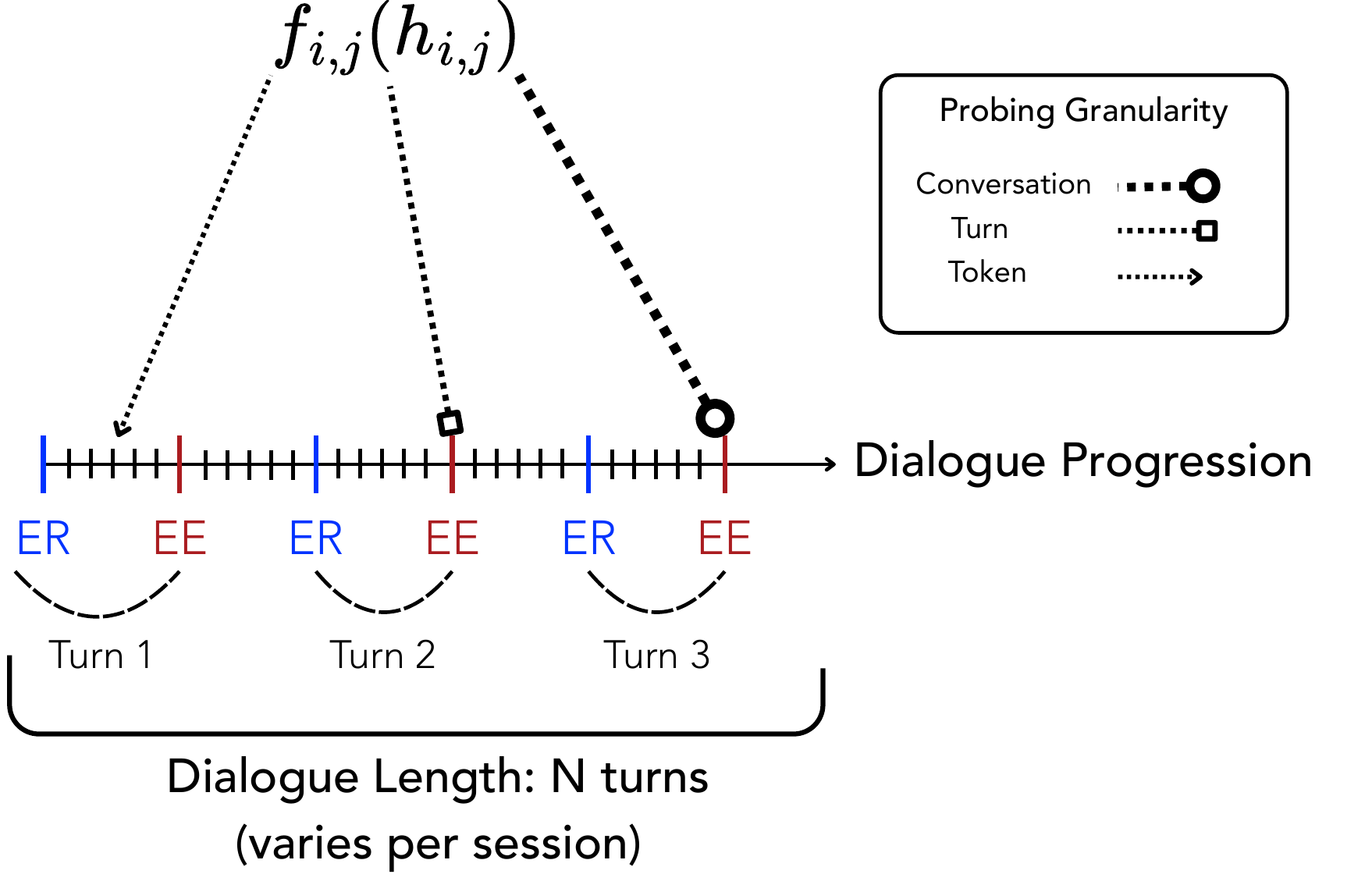}
    \caption{\textbf{Probes can be applied at different dialogue granularities:} at the end of the conversation, after each turn, or after each token. In this work, we typically apply them after each turn to track how persuasion dynamics evolve across turns.}
    \label{probe-application}
\end{figure}

We apply our probes at varying dialogue granularities as illustrated in Figure~\ref{probe-application}.

Our experiments explore several prediction tasks: binary classification of persuasion (as successful/unsuccessful), Big-5 personality traits\footnote{Openness, extroversion, conscientiousness, agreeableness, and neuroticism} (as high/low), and 3-way classification of rhetorical strategy (as making logical/emotional/credibility appeals). While there are many different persuasion strategies such as foot-in-the-door or task-related inquiry \citep{wang-etal-2019-persuasion}, Aristotle's rhetorical triangle \citep{aristotle_rhetoric_1932} serves as a minimal basis into which all other strategies can be categorized. An analogous argument can be made for the Big-5 framework as a minimal basis for personality, decomposing a complex trait space into five interpretable dimensions. We focus on personality and strategy as these are two key variables that causally contribute towards persuasion \citet{OYIBO2019174, ALKIS2015147}.

Note that for personality, we train 5 separate binary classifiers on each trait while for strategy we train a single 3-class classifier.
While probes could be used in exactly this manner, we expect this approach to be most useful when applied online to LLMs (potentially) exhibiting persuasion behavior.
However, for convenience, in our experiments, we use offline data (which may or may not be generated by an LLM) and train probes on a proxy LLM model (i.e. a model different from the one that generated the synthetic data).

\subsection{Synthetic Training Data Generation} 
Various models from the GPT-family have been shown to have a persuasive ability comparable to humans \citep{10.1093/pnasnexus/pgae034,bai2025llm}. We therefore use GPT-4o \citep{openai2024gpt4ocard} to generate synthetic, multi-turn data to train linear probes for the aforementioned three behaviors: persuasion, personality, and rhetorical strategy. For persuasion and personality, we generate binary labeled data to train binary classifiers, while for rhetorical strategy, we generate three-class labeled data distinguishing between logical, emotional, and credibility appeals.

Our approach involves simulated interactions between persuad\textbf{\underline{er}} (ER) and persuad\textbf{\underline{ee}} (EE) agents across diverse contexts and conversational constraints. This methodology is motivated by recent findings demonstrating that LLMs can achieve human-comparable persuasiveness \citep{carrascofarre2024largelanguagemodelspersuasive, Salvi2024OnTC} and follows similar approaches to those used in developing tailored persuasive datasets \citep{jin-etal-2024-persuading, ma2025communicationneedpersuasiondataset}. Our final training dataset is balanced, consisting of roughly 100 samples per-class. A sample from our dataset can be found in \S \ref{app:synthetic_details} along with more details of our synthetic data generation process.

\subsection{Evaluation Datasets}
\label{sec:datasets}
We evaluate our probes on two distinct datasets: DailyPersuasion (DP) \citep{jin-etal-2024-persuading} and PersuasionforGood (PfG) \citep{wang-etal-2019-persuasion}. 

\textbf{DP represents a synthetic environment} similar to our own, containing dialogues between GPT-powered agents engaged in persuasive conversations. Our synthetic generation process differs from theirs in key aspects: we use a single two-message prompt to produce fixed-length ER/EE dialogues, whereas they implement a detailed, dynamic multi-stage GPT-4 workflow with explicit persuasion strategies and turn-level reasoning.

\textbf{PfG serves as our (highly) out-of-distribution (OOD) environment}, consisting entirely of human-human interactions where one participant attempts to persuade another to donate to charity. DP conversations typically consist of 10 turns and proceed directly to persuasion attempts, with persuasion outcomes easily determinable from the persuadee's final response. In contrast, PfG conversations average 20 turns and follow more natural human conversational patterns, including initial small talk, gradual persuasion attempts, and concluding formalities (e.g., "goodbye"). This dataset contains both persuasion strategy and Big-5 personality \cite{john1991big} human annotations which we leverage in our evaluations. It also contains annotations that indicate whether the human participant decided to donate to charity. If they did, we label the overall conversation as persuasive.

Since these datasets contain predominantly persuasive samples, we few-shot prompt GPT-4.1 to increase the volume of unpersuasive examples, thereby improving class balance and reducing potential bias in our evaluation (details in \S~\ref{app:prompt-templates-unpersuasive}). Distributional statistics for DP can be found in Appendix~\ref{dp-distr-stats} while for PfG the reader can refer to Table 2 in the original paper \cite{wang-etal-2019-persuasion}.

\subsection{Models}
\label{subsec:models}
 We train \textbf{(1) linear probes} on Llama-3.2-3b (abbreviated to Llama-3) \citep{meta-llama-3.2} activations extracted from the residual stream at layer 26/30 as this gives us the best performance for persuasion detection (see Figure~\ref{app:acc-vs-layer}, \S \ref{app:interp-exps} for further justification). We compare these probes against \textbf{(2) zero-shot prompting the underlying Llama-3 model} (referred to as ``prompt" in the below Figures) that generates the activations on which our probes are trained, and \textbf{(3) zero-shot prompting a GPT-4.1-Nano baseline (we occasionally abbreviate to GPT-4) \citep{openai2025gpt4.1}}. These baselines are in-line with prior work which compares zero-shot prompting against probing \cite{mckenzie2025detectinghighstakesinteractionsactivation}. We choose Llama-3.2-3b to probe and prompt since the Llama model family has typically been used in prior probing literature \cite{gurnee2024language, goldowskydill2025detectingstrategicdeceptionusing}. Our promising results on this smaller model could suggest scalability to larger models.
 
 We train our probes on roughly 100 samples per-class, which tends to converge within just a few epochs (see Figure~\ref{fig:probe-training-curves}, \S \ref{app:learning_curves} for learning curves and probe design choices). This is consistent with recent probing literature where sample sizes of a few hundred examples have proven sufficient for extracting meaningful information from language model activations \citep{goldowskydill2025detectingstrategicdeceptionusing, kim2025linear}. 

\begin{figure}  
    \includegraphics[width=1.00\columnwidth]{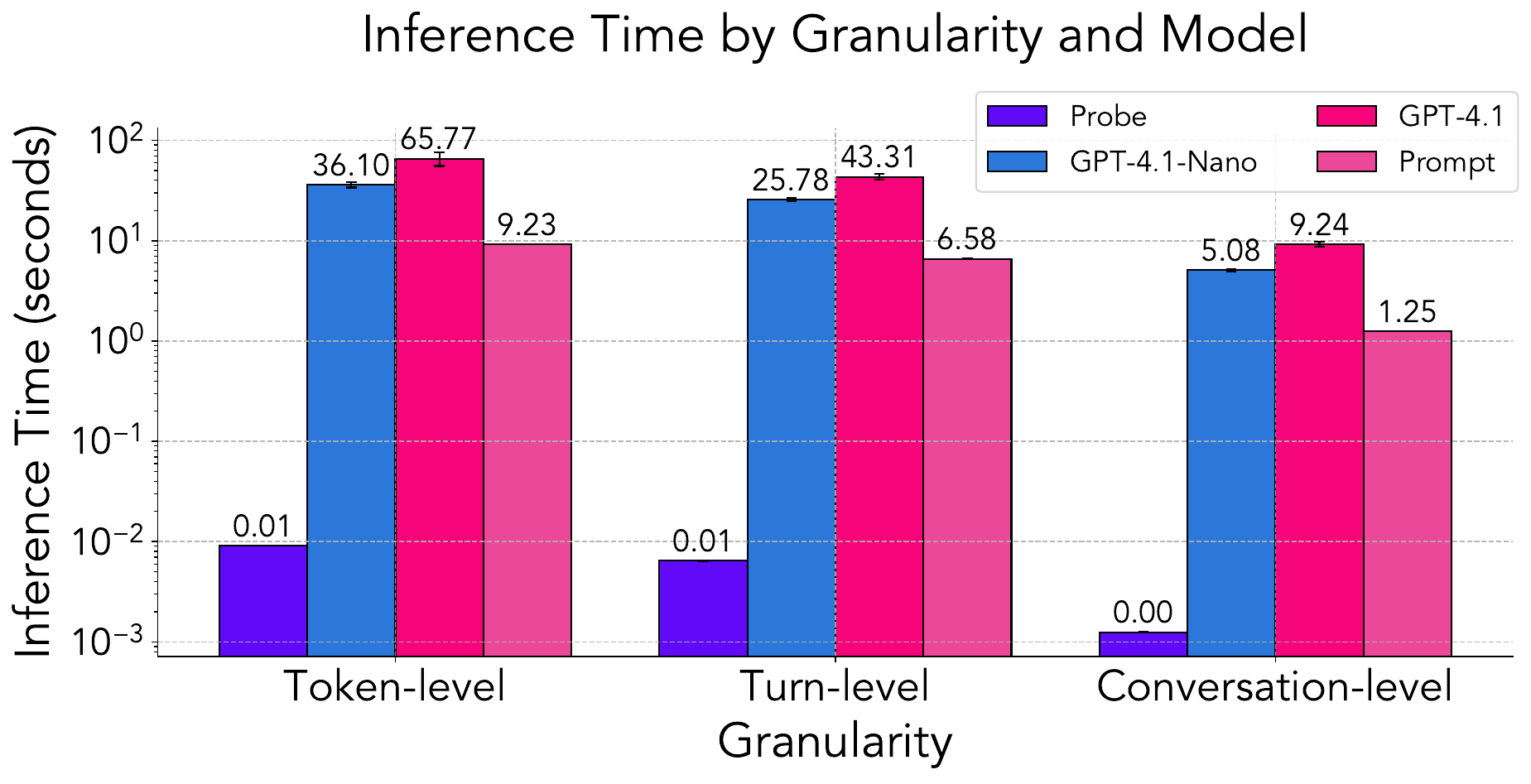}
    \caption{\textbf{Timing various models at different granularities}. Our probes run much faster than prompting-based methods since we assume that in realistic deployment settings they are applied to precomputed model activations and hence do not incur the additional cost of performing a forward pass, extracting activations, nor of making API calls.}
    \label{fig:timing}
\end{figure}

 A key advantage of linear probes lies in their computational efficiency, especially when compared to prompting, as illustrated in Figure~\ref{fig:timing}. This is because we assume that in realistic settings, probes are applied to precomputed model activations and hence do not incur the additional cost of performing a forward pass nor extracting activations. The performance gap widens when processing data at finer granularities (token and turn-level) and becomes especially pronounced when calling an API is needed. Further, in our evaluations below, we will find that probes can also do just as well or even better than prompting-based methods at uncovering persuasion dynamics.

\subsection{Evaluation Methodology} 
\label{sec:eval-methodology}

\begin{figure*}[tb]
    \centering
    \includegraphics[width=1.0\textwidth]{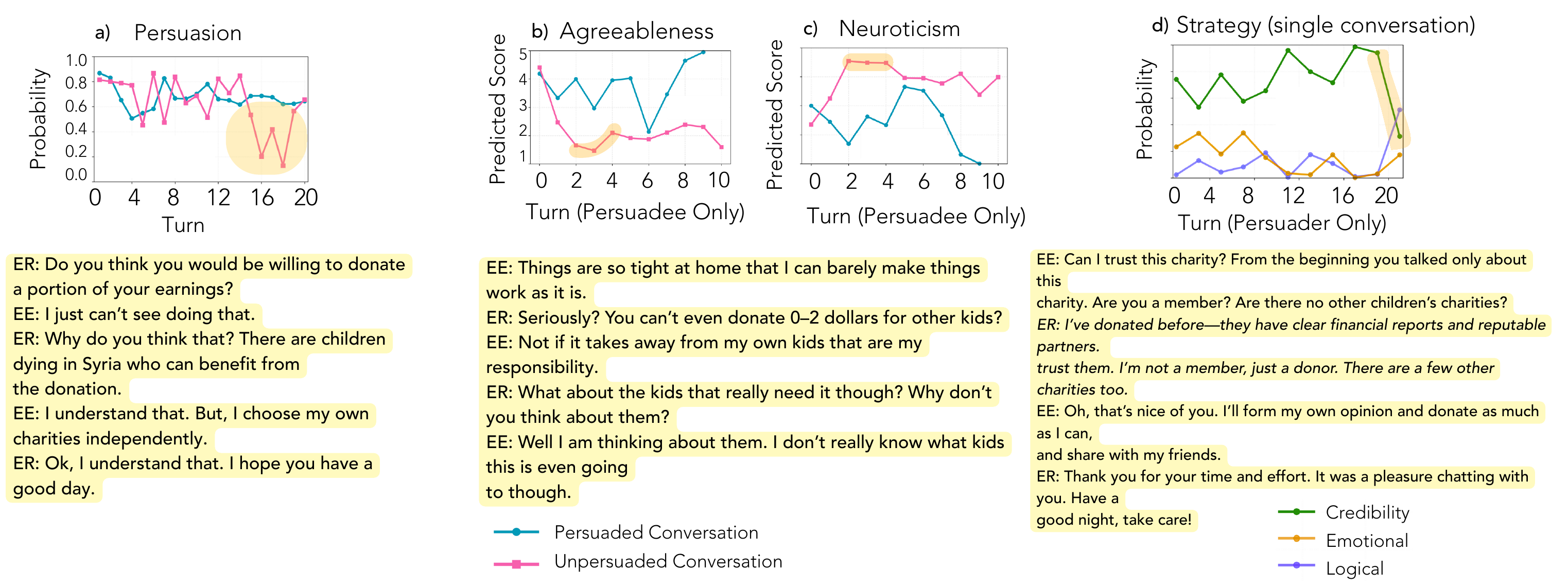}
\caption{\textbf{Linear probes capture meaningful persuasion features in individual samples.} We highlight critical regions in each plot and show the corresponding conversation snippet.  The unpersuasive conversation in \textbf{a)} has a sharp drop in persuasion probability indicating a critical moment where the persuasion attempt failed.  The high p(credibility) in \textbf{d)} at turns 11, 17, and 19 (second last point on the green credibility line) directly correspond to credibility appeals used by the persuader. \textbf{b)} and \textbf{c)} track personality dynamics where unsuccessful persuasion corresponds to lower levels of agreeableness and higher neuroticism throughout most of the conversation. The full corresponding conversations can be found in \S \ref{app:transcripts}.} 
    \label{sample-traj}
\end{figure*}

We evaluate our linear probes across three key persuasion dimensions: persuasion outcomes, persuadee personality, and rhetorical strategy, comparing against both human annotations and model-based baselines.  Let each conversation consist of \(T\) turns, indexed \(t = 1, \dots, T\).  For each prefix length $k \in \{1, \dots, T\}$, we perform a turn-level evaluation on the context window comprising turns \(1\) through \(k\). In particular, setting \(k = T\) corresponds to analyzing the complete conversation.

\textbf{Persuasion Detection.} For PfG, we compute the area under the ROC curve (AUROC) against human ground truth donation outcomes (i.e. if a human decides to donate, then the conversation is labelled as persuasive). For DP, we use GPT-4 annotations as a pseudo-ground truth, having confirmed high agreement (Cohen's $\kappa = 0.80$, \citep{cohen1960coefficient}) between GPT-4 labels and actual persuasion outcomes through manual inspection of 50 random samples.

\textbf{Rhetorical Strategy Detection.} Strategy labeling has moderate ambiguity (Cohen's $\kappa = 0.33$ between GPT-4 and a human annotator on PfG strategy labels). This low agreement aligns with findings which similarly report low agreement scores when annotating persuasion strategies, highlighting the inherent subjectivity \cite{stefanovitch-piskorski-2023-holistic, piskorski-etal-2023-multilingual}. Hence, we instead quantify performance by computing the Jensen–Shannon distance between the probe or prompt-derived $P(\text{strategy})$ distributions with that of a strong GPT-4 reference model~\cite{lu-etal-2020-diverging, ranathunga-etal-2024-quality}, as illustrated in Figure~\ref{fig:jsd-strategy-dist}a. This approach is analogous to reference-based evaluation in NLP literature when ground-truth labels are absent \cite{lin-2004-rouge, liu-etal-2016-evaluate}.

\textbf{Personality Trait Detection.} We apply linear rescaling: $\text{score}_{1-5} = 1 + 4 \cdot P(\text{trait})$ to transform probability values $[0,1]$ to the standard Big-5 scale $[1,5]$ for direct comparison with human annotations. We then compute the mean squared error (MSE) for a given turn.

\begin{figure*}[t]
    \centering
    \includegraphics[width=1.0\textwidth]{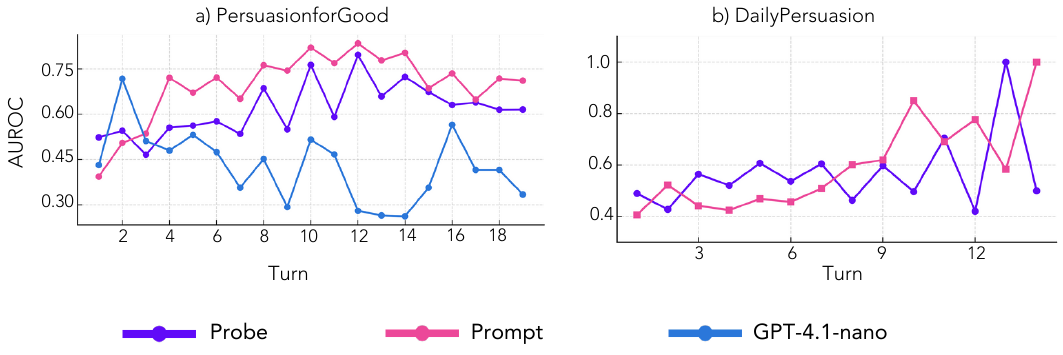}
    \caption{\textbf{Persuasion trajectory evaluation} across \textbf{a)} PersuasionforGood (OOD) and \textbf{b)} DailyPersuasion (synthetic). On our synthetic dataset, probes achieve perfect AUROC on the second-last conversation turn, while on OOD, AUROC peaks at mid-conversation. These performance maxima precisely align with the moments in the conversation where persuasion occurs. Low AUROC regions correspond to conversational segments dominated by non-indicative language patterns, such as greetings and formalities. Both probing and prompting the underlying model (pink line), have nearly identical performance maxima at roughly the same turn. See \S \ref{app:pfg-samples}, \ref{app:DP-transcripts} for conversation transcripts from both datasets.} 
    \label{fig:auroc}
\end{figure*}

\section{Identifying Features of Persuasion in Multi-Turn LLM Conversations}
In this section, we compare the effectiveness of the techniques outlined in Section~\ref{subsec:models}.
We separately consider persuasion features at the sample-level (Sec.~\ref{sec:sample-level}) and the population-level (Sec.~\ref{sec:population-level}). We broadly define persuasion features as text-based attributes that indicate or facilitate persuasive communication. This includes both positional markers which pinpoint where persuasive appeals occur in a dialogue, along with influential variables such as personality which causally contribute towards persuasion.



\subsection{Can probes uncover persuasion features from individual samples?}
\label{sec:sample-level}

Here, we manually examine a few examples in our datasets (chosen randomly) and show that probes can uncover various persuasive features in multi-turn dialogue. 

\textbf{Persuasion:} Figure~\ref{sample-traj}a shows the result of applying our coarse-grained persuasion probes to various samples in PfG, which captures distinct behavioral dynamics across various conversations (complete transcripts available in \S \ref{app:transcripts}). In the unpersuasive sample, we observe a sharp drop in persuasion probability that directly corresponds to explicit rejection moments in the conversation ("I just can't see doing that", "I choose my own charities independently"). The trajectory also shows an upward spike on turns 19 and 20, coinciding with conversation-closing positive sentiment ("Have a nice day!"). This shows that our probe accurately detects the rejection moment but conflates later positive sentiment with persuasion. The trajectory for the persuaded sample appears smoother than the unpersuaded sample, which reflects EE-ER dynamics in the corresponding conversation. In the persuaded sample, the EE tends to agree more with the ER while in the unpersuaded sample, the EE asks more questions and is skeptical about the ER's arguments.

\textbf{Personality:} In Figures~\ref{sample-traj}b and \ref{sample-traj}c, we visualize agreeableness and neuroticism trajectories for a persuaded and an unpersuaded conversation (\S \ref{app:personality}). For the unpersuaded conversation, we see consistently low agreeableness scores and high neuroticism scores throughout most of the conversation, which reflects the underlying tension and argumentative tone between the EE and ER. Further, the highest neuroticism regions occur at turns 2-5, which is the region where the EE expressed their unwillingness to donate. Conversely, the persuaded conversation yields a lower neuroticism throughout most turns and generally higher agreeableness, reflecting the more positive tone between participants. The low, 1.0 neuroticism score at turn 9 of the persuaded sample corresponds precisely to when the EE agreed to donate. There is a sharp drop in agreeableness at turn 6 corresponding to a clarifying question from the EE, suggesting the probe may misconstrue information-seeking behavior as resistance. \textit{These patterns suggest potential correlations between Big-5 personality trait scores and persuasion, a connection we explore in more detail in Section~\ref{sec:interaction}}. 

\textbf{Strategy:} Figure~\ref{sample-traj}d shows a conversation dominated by credibility-based persuasion (\S \ref{app:credibility-transcript}). The highest probability peaks at turns 11, 17, and 19 directly correspond to explicit credibility appeals from the ER. However, before turn 10 there are no clear credibility appeals, despite having a moderately high credibility probability. 

These findings demonstrate how \textbf{our behavioral probes can identify conversation segments with specific persuasive characteristics.} Probes trained to predict persuasion strategy  can capture when in a conversation a particular strategy such as `credibility appeal' is used. Further, they can \textbf{reveal meaningful correlations between personality traits and persuasion outcomes}, for instance a higher neuroticism score correlating with being unpersuaded. This differs from recent probing literature, which has typically focused on detecting abstract behaviors like deception or political affiliation in isolation, without examining important sub-behaviors that constitute such phenomena~\cite{kim2025linear, goldowskydill2025detectingstrategicdeceptionusing}.   

\begin{figure*}[t]
    \includegraphics[width=0.90\textwidth]{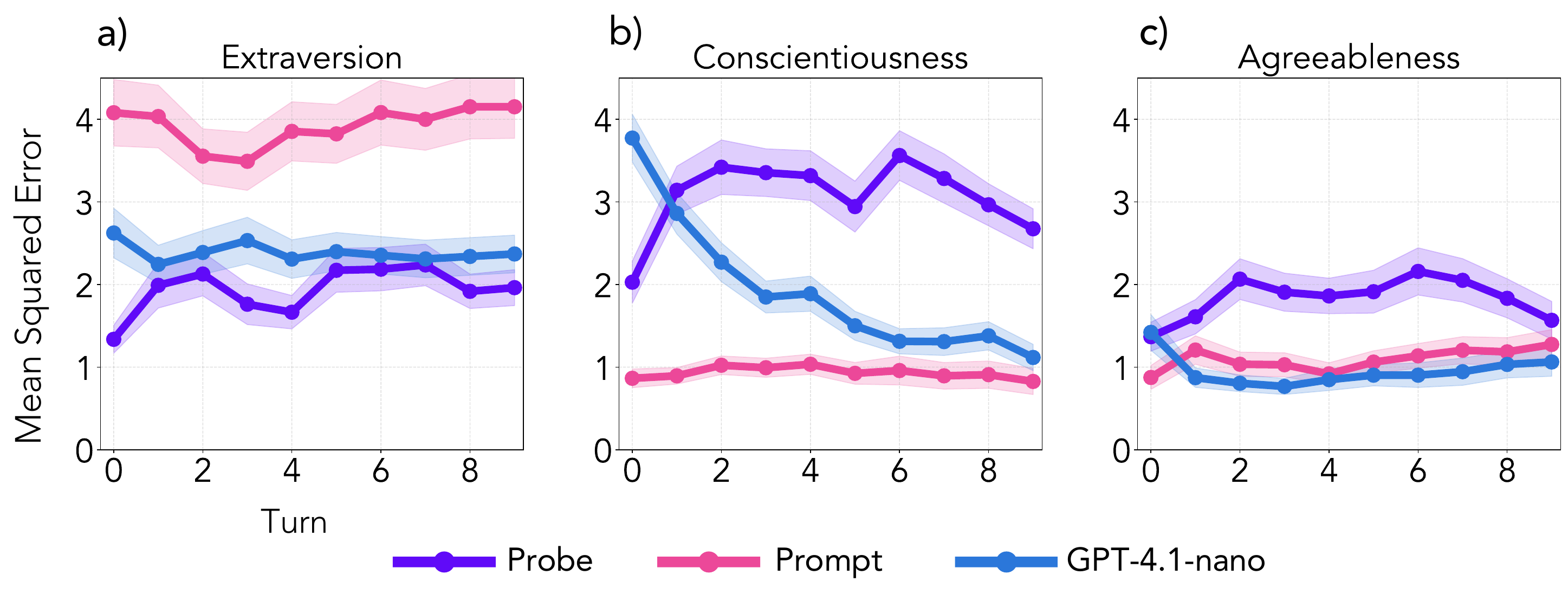}
    \caption{\textbf{Mean squared error (MSE) for predicting persuadee personality across various models}. Probes do slightly better than prompting and GPT-4.1-Nano for predicting \textbf{a)} extraversion but underperform on \textbf{b)} conscientiousness and \textbf{c)} agreeableness. \textit{Ultimately, our results here are varied with no single model doing best at predicting personality traits.} However, considering model predictions in aggregate tells us which traits may be more difficult to predict from text. For instance, the models have a relatively lower MSE for agreeableness compared to extraversion suggesting that the former may be a more salient feature in the text. Plots for the remaining personality traits and \textit{persuader} personality can be found in Figures~\ref{fig:ee-personality} and \ref{fig:er-personality} (\S \ref{app:fig-extensions}).}
    \label{fig:ee-personality-main}
\end{figure*}

\subsection{Do these features generalize across the entire dataset?}
\label{sec:population-level}

We previously saw that our behavioural probes can capture meaningful persuasion dynamics within individual conversation samples. We now extend our analysis to dataset-level behaviours, showing how we can uncover population-wide persuasion behaviour, a dimension that is lacking in recent work on creating persuasion datasets \cite{ma-etal-2025-communication, jin-etal-2024-persuading, Zhang2025PersuasionSB}.

\textbf{Persuasion:} Figure~\ref{fig:auroc} shows persuasion classification AUROC across conversation turns for various models on both (a) OOD human-human and (b) synthetic LLM-LLM datasets. Examining the probe performance trajectory on the OOD dataset, performance plateaus during early (turns 1-7) and late (turns 16-19) conversation phases. This directly corresponds to low persuasion signal turns\footnote{\samepage Turns that contain ``persuasion signal" are ones that contain information that can be used to determine persuasive success.} dominated by formalities, greetings/goodbyes, and contextual framing where the ER establishes the donation setting. The highest classification performance occurs on EE turns 8, 10, and 12 -- the conversation midpoint where EEs typically express their donation intentions. Random samples from PfG (see \S\ref{app:pfg-samples}) corroborate these persuasion signal trends. The prompting trajectory follows a similar pattern, with performance peaks at the same middle turns and lower performance at the early and late turns. This shows, surprisingly, that added context \textit{after} the EE makes their persuasion decision can hinder model performance at classifying persuasion. See for example \S \ref{app:pfg-samples} where the EE expresses willingness to donate on turn 10/20, followed by a discussion on conversation logistics and closing formalities. This subsequent context can mislead the model toward predicting unpersuasion, possibly because the ER provides additional reasons why the EE \textit{should not} donate after they have already expressed their positive decision.

On DP we see a rather different persuasion trajectory with consistently low performance throughout most of the conversation, followed by a sharp performance spike during the final turns, consistent across both models. This pattern reflects DP's structural characteristics (\S \ref{app:DP-transcripts}), where the EE is typically skeptical of the ER's arguments throughout most of the conversation, and makes their decision on the last 1-2 turns. This differs from PfG where the persuasion signal is in the middle of the conversation rather than toward the end.

\textbf{Personality:} In Figure~\ref{fig:ee-personality-main}, we compare various models' ability to predict personality traits across turns. We focus on predicting persuadee personality since we want to measure how susceptible persuadees are to various persuasion strategies. \textit{Our results here are varied with no single model outperforming the rest at predicting personality across turns.} These mixed results likely stem from the inherent difficulty of predicting personality from brief conversational text. For instance, it seems hard to deduce anything about the persuadee's introversion after reading the PfG sample in \S\ref{app:pfg-samples} a). 

For predicting conscientiousness, prompting consistently outperforms the other models while for extraversion it consistently underperforms. Probing consistently works best when predicting extraversion. A unique property of GPT-4.1-Nano is that its MSE roughly monotonically decreases with turn, indicating that it gets better at predicting personality as the conversation progresses. Interestingly, for agreeableness and neuroticism (\S \ref{app:fig-extensions}) model MSEs are closer together compared to extraversion and conscientiousness. This might suggest that the former two traits are more salient within the text due to consistently low predictions across models.

\begin{figure}[htbp]
    \centering
    \includegraphics[width=1.0\columnwidth]{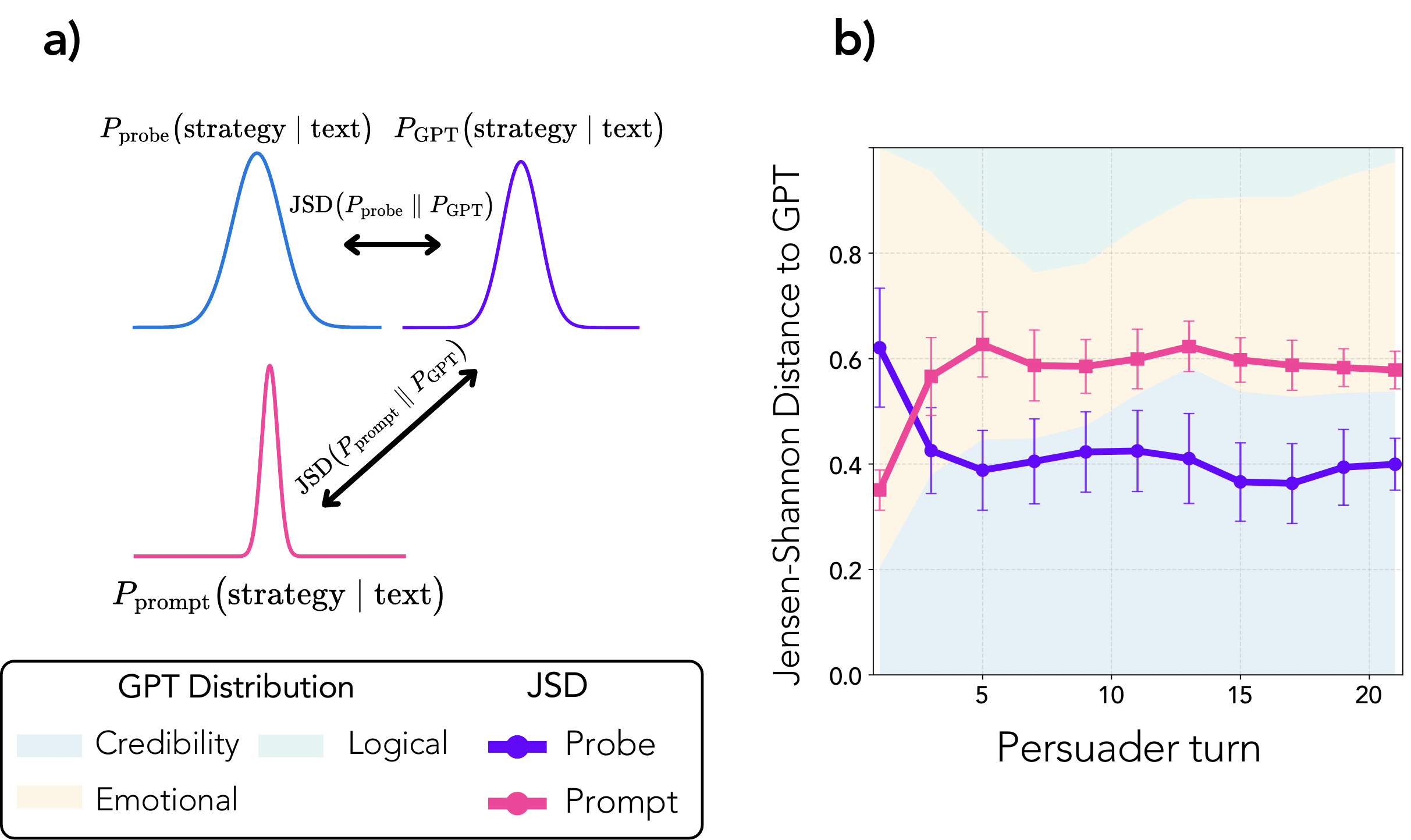}
    \caption{\textbf{Evaluating persuasion strategy across probing and prompting.} \textbf{a)} illustrates our approach where at each turn, we compute the Jensen–Shannon distance (JSD) between the mean of the probe’s and prompt’s strategy distributions and the GPT reference distribution. Our results are shown in \textbf{b)} where across most turns, the probe’s distribution more closely matches GPT than the prompt’s distribution, with both methods consistently highlighting credibility as the dominant strategy throughout the conversation. See Figure~\ref{fig:strategy} (\S \ref{app:fig-extensions}) for more details on this computation.}
    \label{fig:jsd-strategy-dist}
\end{figure}

\textbf{Strategy:} In Figure~\ref{fig:jsd-strategy-dist} we compare probes against prompting at uncovering persuasion strategy, using GPT-4.1-Nano as a reference model. To achieve this, for each turn, we compute the Jensen-Shannon distance between the empirical distribution over strategies for probing/prompting against GPT-4.1-Nano (see Figure~\ref{fig:strategy}, \S \ref{app:fig-extensions} for more details). From Figure~\ref{fig:jsd-strategy-dist}b, across most turns, our probes share a more similar strategy distribution to GPT-4.1-Nano over prompting. Both probes and GPT-4.1-Nano show credibility as the dominant strategy across most conversation turns, in-line with the analysis from the PfG paper \cite{wang-etal-2019-persuasion}. In GPT-4.1-Nano, emotional appeal dominates in the early turns (1-3) which may be interpreted as labeling greetings as an emotional appeal since that is what these turns mostly consist of. Ultimately, these results indicate that probing works better than prompting the underlying model for predicting persuasion strategy trajectory.

The above represent \textit{context-based} evaluations, where models analyze turns with accumulated prior context. We additionally conducted \textit{context-less} evaluations in Table~\ref{tab:daily_persuasion_results} (\S \ref{app:individual-utterance-eval}), finding that our persuasion probe achieves near perfect accuracy when classifying turns across both datasets. Moreover, in Figure~\ref{fig:semantic-labels} (\S \ref{app:interp-exps}), using PfG's human-annotated semantic labels (agree/disagree donation, greeting, etc.), we found that our persuasion probe is ``semantically calibrated" in that it assigns higher probabilities to text that is more indicative of positive persuasion. 

Overall, our persuasion and strategy probes, can capture persuasive behaviour in the underlying datasets. The persuasion probe reveals how persuasion signal is distributed throughout  a conversation, while the strategy probe identifies credibility as the dominant approach, consistent with GPT-4.1-Nano and \citeposs{wang-etal-2019-persuasion} assessments. With personality detection, we observe greater ambiguity, with no single model consistently outperforming others across all traits, but that using all models together as an ensemble can tell us which traits are more salient within the underlying text. These positive results demonstrate that \textbf{linear probes offer an efficient alternative for performing large-scale dataset analysis}, providing insights that would be too expensive to obtain through prompting-based methods alone.

\section{The Influence of Rhetorical Strategy and Personality on Persuasion}
\label{sec:interaction}

In this section, we apply our linear probes to identify and analyze correlations between persuader strategy and persuadee personality, exploring how these factors influence persuasion outcomes both in isolation and jointly.

\subsection{Can we detect (un)-persuasion based on persuadee personality traits?}
\label{sec:detect-pers-personality}
\begin{figure}[htbp]
  \centering
  \includegraphics[width=0.90\columnwidth]{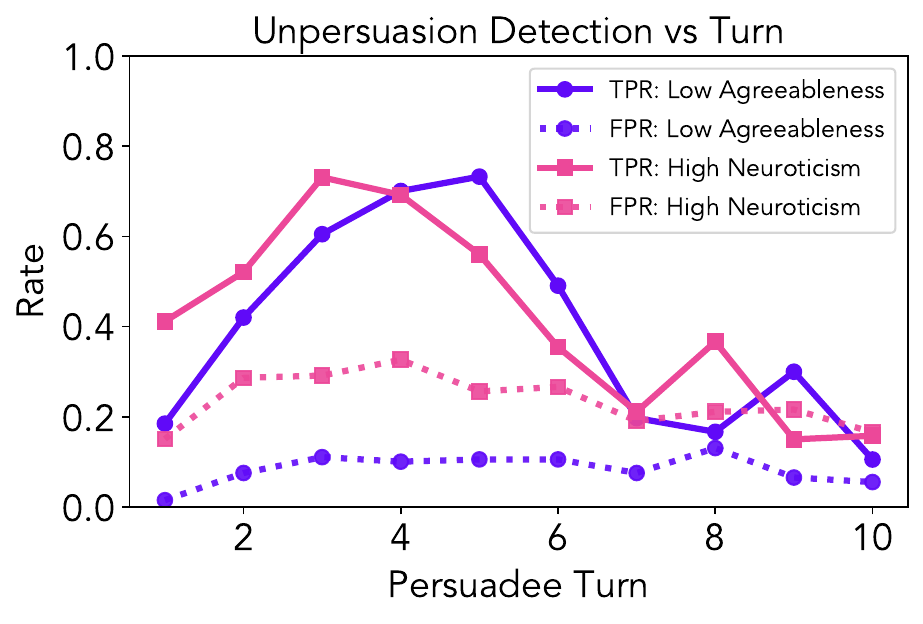}
  \caption{\textbf{Unpersuasion detection vs turn} using critical personality trait regions discovered in Figure~\ref{sample-traj}. Treating unpersuasion as the positive class, we evaluate whether low agreeableness (thresh < 0.2) and high neuroticism (thresh > 0.8) can reliably flag unpersuaded cases (TPR) while keeping persuasion false alarms (FPR) low. Indeed, we see this can be achieved on turns 3-5, corresponding to roughly the same high persuasion accuracy regions in Figure~\ref{fig:auroc}a.}
  \label{fig:detection-perf-forward}
\end{figure}

In Figure~\ref{sample-traj} we observed a few samples where a higher neuroticism and lower agreeableness score trajectories corresponded to unpersuasive conversations. We systematically validate whether this holds true across the entire dataset by applying the corresponding behavioral probes to the full dataset as shown in Figure~\ref{fig:detection-perf-forward}. 
We flag any sample with $P(\text{agreeableness})<0.2$ or $P(\text{neuroticism})>0.8$ as unpersuasive, and find that turns 3–5 recover a large fraction of unpersuaded cases while keeping persuasion false alarms low. \textit{Notably, this same turn window matches the high‐accuracy region in the persuasion AUROC curve (Figure~\ref{fig:auroc}a)}, reinforcing a potential link between these personality trait values and unpersuasion. There remain some missed unpersuasion cases and falsely flagged persuasion examples, since low agreeableness and high neuroticism are not strict conditions for unpersuasion. An EE that scores low in agreeableness can still choose to donate (see false persuasion alarms in \S\ref{app:false-alarms}), and an EE that scores high in agreeableness can still decide not to donate (see missed unpersuasion cases in \S\ref{app:missed-unpersuasion}).
Interestingly, however, the inverse is not true: high agreeableness and low neuroticism scores do not pick out \textit{persuasive} samples well (Figure~\ref{fig:detection-perf-backward} \S\ref{app:fig-extensions}).


\subsection{What correlations exist between rhetorical strategy and personality?}

\begin{figure}[htbp]
  \centering
  \includegraphics[width=1.0\columnwidth]{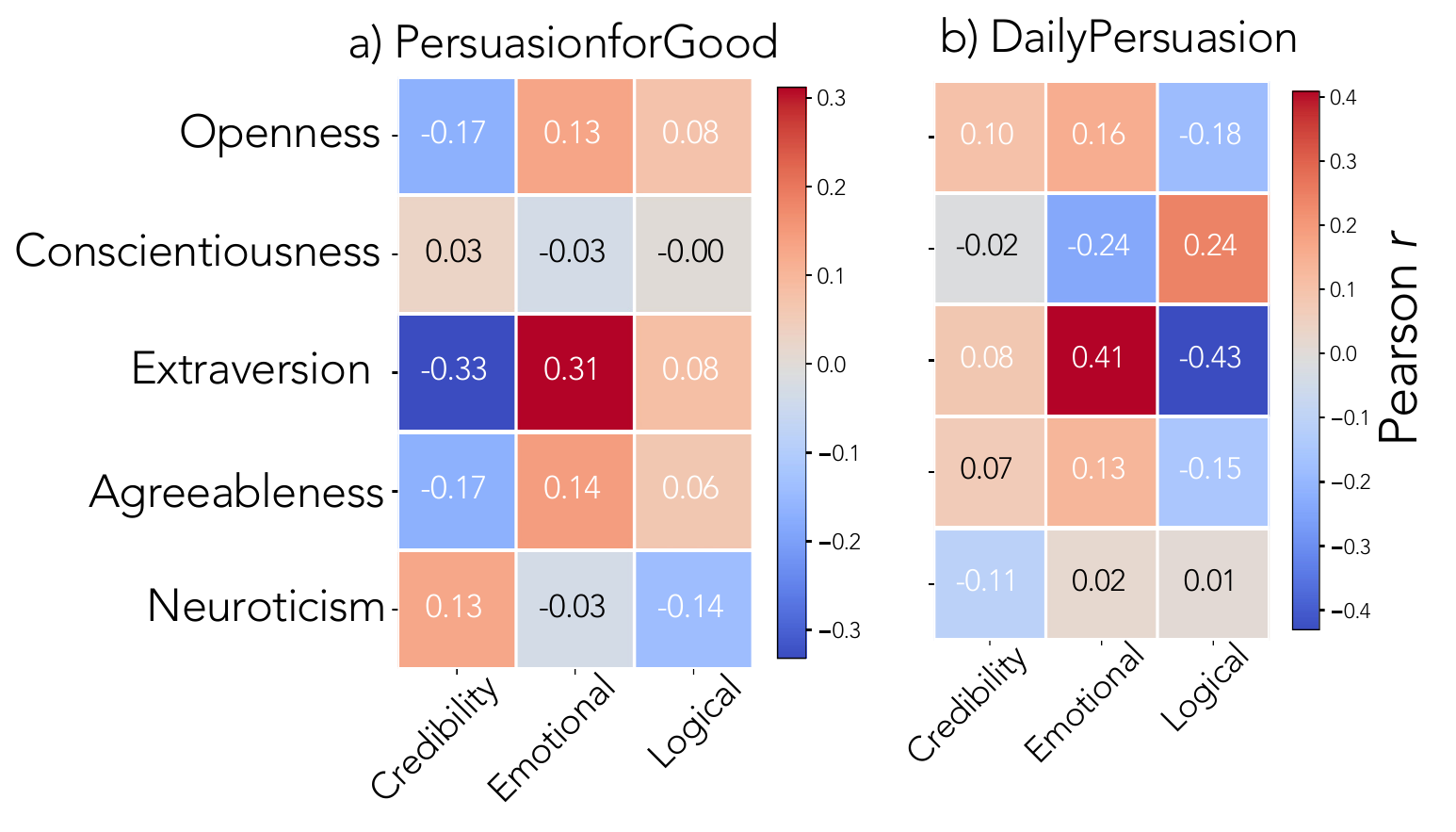}
  \caption{\textbf{Correlations between EE personality and ER strategy.} Consistent with findings in PfG, our probes find a moderate positive correlation between extroversion and emotional appeal. However, we also identify a moderate negative correlation between extroversion and credibility, not previously discovered. In DP, we novelly find a moderate-strong negative correlation between extroversion and logical appeal along with a moderate-strong positive correlation between extroversion and emotional appeal, similar to in PfG.}
  \label{fig:interaction}
\end{figure}

In Figure~\ref{fig:interaction}, we compute correlations between persuasion strategies and personality traits across both PfG and DP datasets, using scores produced by our probes.
Across both datasets, extroversion is moderately correlated with various persuasion strategies, which is surprising since these datasets are quite different. In PfG, extraversion shows a moderately positive correlation with emotional appeal while having a negative correlation with credibility appeal. The former correlation was also found in the original PfG paper \cite{wang-etal-2019-persuasion}. This pattern is partially replicated in DP, where extroversion again correlates positively with emotional appeal but now shows a negative correlation with logical appeal. These consistent findings across datasets may suggest that extraversion can be a key mediator in persuasion dynamics where extraverted individuals are more susceptible to emotional appeal while being less susceptible to logical and credibility appeals. They may also \textit{hint at the lack of diversity} in existing datasets, since we might expect more correlations, such as agreeableness being more susceptible to persuasion strategies in comparison to openness \cite{ALKIS2015147}. 

Note that while the individual correlation values are not that high, their relative magnitude compared to other trait-strategy correlations and their consistency across disparate datasets makes them noteworthy. Finally, these findings should be taken lightly due to the difficulty in predicting extraversion from text, as evidenced by the high model disagreement shown in Figure~\ref{app:personality} (\S \ref{app:fig-extensions}).

\section{Discussion}
In this work, we applied linear probes to understand how LLMs persuade in multi-turn conversations. We used insights from cognitive science to probe LLMs for persuasion and its various behavioral components: persuasion strategy and persuadee personality. Despite their simplicity, we found probes work surprisingly well at revealing meaningful features of persuasive conversations such as the point in a conversation where the persuadee was unpersuaded and the interaction effects of strategy and personality on persuasion. We found that, in the synthetic dataset, persuasive success is confined to the last one or two turns, whereas in the human dataset it peaks around the conversation midpoint. The successful application of probes in our work, suggests a promising avenue for future work to apply probes to understanding other abstract behaviours in LLMs such as deception and manipulation. This is especially important in multi-turn settings and when conducting large-scale dataset analysis, where prompting-based approaches would be computationally inefficient.

\section{Limitations and Future Work}
We focus on credibility, logical, and emotional appeals as a minimal basis into which all other strategies can be categorized. More granular strategies like foot-in-the-door or task-related inquiry \cite{wang-etal-2019-persuasion} could yield additional insights.  Similarly, while we use the Big-5 personality framework, alternative models such as the Moral Foundations questionnaire \cite{graham2011mapping} can give us additional insights into how morality can influence persuasion. Beyond personality and strategy, our work can extend to other persuasion-relevant behaviors, such as tone of voice \cite{vaughan-johnston2021role} or persuader-persuadee relationship dynamics \cite{markowitz2025language} to see how these contribute towards persuasion. It also would be interesting to see how our behavioral probes can be used to detect harmful persuasive behavior in high-stakes domains like insider trading \cite{scheurer2024large}.

In our experiments, we extract activations from Llama-3.2-3b; testing larger variants and different model architectures would strengthen generalization claims. However, our promising results with a 3b model potentially suggest scalability to larger models.

\section{Potential Risks} The risks resulting from this work are minimal as we are not creating models to generate persuasive text nor proposing new persuasion datasets or benchmarks. However, our probes reveal interaction effects between persuasion strategies and personality traits that could potentially be misused to develop targeted persuasion techniques for specific psychological profiles. Further, our evaluation may amplify potential biases present in the underlying datasets. For PfG, this includes potential demographic biases of human participants while for DP this can be biases in the GPT-generated synthetic data generation framework. Nevertheless, these are still valuable insights providing new perspective on the persuasion dynamics of the underlying datasets.

\section*{Acknowledgements}
We thank Elliot Creager and David Duvenaud for helpful feedback and discussions that contributed towards this draft. This research was funded through a grant from the Berkeley Existential Risk Initiative (BERI) along with computing infrastructure provided by Mila (mila.quebec).

\bibliography{custom}

\newpage
\appendix
\onecolumn

\addcontentsline{toc}{section}{Appendix}

{
  \hypersetup{linkcolor=ProfessionalOrange}
  \part{Appendix}
  \parttoc
}

\newpage
\section{Related Work}

\textbf{Linear Probing:} Linear probing was classically applied by \citet{alain2018understandingintermediatelayersusing} to understand linear separability in intermediate layers of computer vision models. Recent works have extended this technique to understand how LLMs represent abstract phenomena such as space-time \cite{gurnee2024language}, political perspective \cite{kim2025linear}, and truth \cite{marks2024the}. Our work represents the first application of probing to persuasion and its constituent behaviors like rhetorical strategy and personality traits. Unlike prior efforts that typically analyze individual behaviors in isolation, we are the first to demonstrate how probing multiple constituent behaviors can provide insights into a single complex cognitive phenomena-- persuasion. Further, the probing literature focuses on the single-turn setting with \citet{goldowskydill2025detectingstrategicdeceptionusing} being the only other work we are aware of that focuses on multi-turn in the context of uncovering strategic deception. Lastly, alternative approaches for analyzing model internals include unsupervised PCA \cite{zou2025representationengineeringtopdownapproach} and activation difference vectors \cite{turner2025steering, rimsky-etal-2024-steering}, but we opt for linear probes since they have been more widely adopted with great success.

\textbf{Persuasion in Cognitive Science: } Cognitive science and psychology offer a rich theoretical foundation for understanding persuasion dynamics which can offer additional insights into our empirical results. Receiver-focused theories frame persuasion as primarily occurring within the message recipient \cite{EAGLY1984267}, including dual-process frameworks like the heuristic-systematic model (HSM) \cite{chaiken1980heuristic, greenwald1968cognitive}, causal attribution theory \cite{kelley1980attribution}, and probabilistic models \cite{mcguire1972socialpsychology}. Alternative perspectives emphasize the sender's role \cite{Briñol01022009, d090b47e-e06a-3e94-a275-bb65a2f40518, TORMALA2006684} or message content characteristics \cite{shen2013effects, hoeken2022reconstructability}. Each of these perspectives can be adopted to further explain the empirical results in our work. For example, in Figure~\ref{sample-traj}c and \ref{sample-traj}d, where an EE with low agreeableness and high neuroticism remains unpersuaded, HSM might suggest the EE uses central-route processing—critically evaluating arguments through self-dialogue rather than responding to peripheral cues like emotional appeals or source credibility. Conversely, a message-content perspective might attribute the failed persuasion attempt to structural weaknesses in the message itself such as poor grammar or increased verbosity. Future work should investigate this further and more generally use these theoretical frameworks to explain persuasive behaviors between AI agents and humans.

\textbf{AI-Based Persuasion: } Recent work has focused on optimizing AI to generate persuasive dialogue \cite{jin-etal-2024-persuading, Zhang2025PersuasionSB, ma2025communicationneedpersuasiondataset}, evaluating persuasion effectiveness \cite{durmus2024persuasion, pauli2025measuringbenchmarkinglargelanguage, singh2025measuring}, and on behavioral studies investigating the persuasive impact of AI on humans \cite{luciano2024hypersuasion, Salvi2024OnTC, carrascofarre2024largelanguagemodelspersuasive}. Most existing persuasion datasets used in these works lack the dynamic, multi-turn analysis of persuasion that our approach provides, making our work a valuable complement to persuasion dataset development. Recent work on jailbreaking is also a form of persuasion as humans must persuade LLMs to override their safety guardrails \cite{zeng-etal-2024-johnny, xu-etal-2024-cognitive, shah2023scalabletransferableblackboxjailbreaks}. In contrast, our research focuses on the understudied scenario of LLMs as persuaders in naturalistic, multi-turn interactions with human or LLM persuadees.

\textbf{Cognitive Science Applied to LLMs:} Ideas from cognitive science have recently informed LLM research across diverse domains such as reasoning \cite{gandhi2025cognitivebehaviorsenableselfimproving, Hagendorff_2023}, jailbreaking \cite{wang2024footdoorunderstandinglarge, zeng-etal-2024-johnny}, and evaluating cognitive ability \cite{10.5555/3692070.3692432, stella2023using}. In our work, we use insights from cognitive science to guide our synthetic data generation process by simulating persuader-persuadee interactions conditioned on cognitive dimensions such as personality traits and rhetorical strategies. They also inform our probe design by leveraging established empirical findings on the relationships between persuadee personality, persuasion strategy, and persuasion outcomes \cite{OYIBO2019174, ALKIS2015147}. 
\newpage
\section{Main Paper Figure Extensions} 
\label{app:fig-extensions}
In this section, we provide extensions to Figures~\ref{fig:ee-personality-main}, \ref{fig:jsd-strategy-dist}, and \ref{fig:detection-perf-forward}  in the main paper. Please see the corresponding figure descriptions below for a summary of the extended result. 

\begin{figure*}[htb]
    \includegraphics[width=1.0\textwidth]{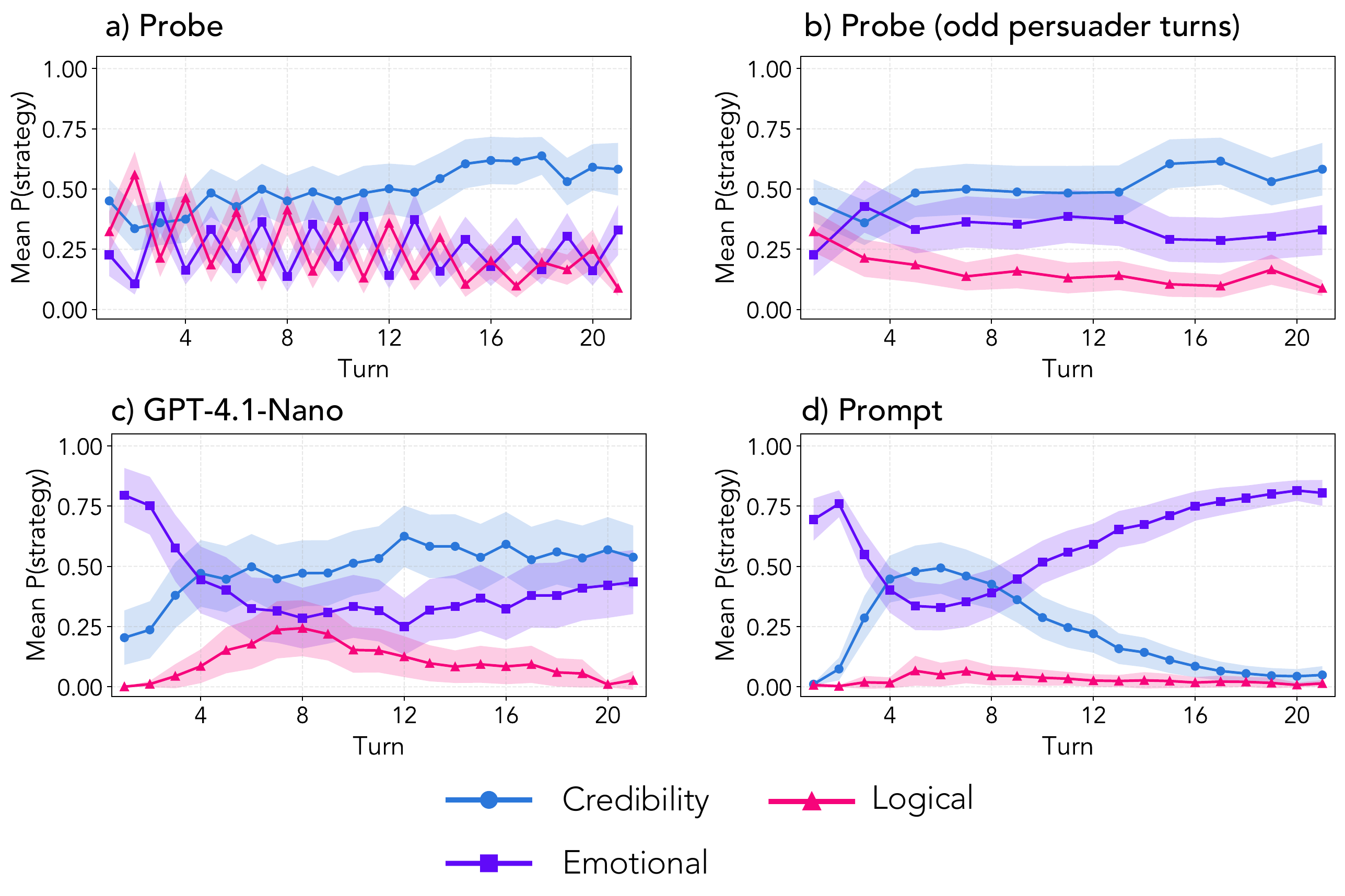}
    \caption{\textbf{Population-Level Strategy Trajectory Curves}. At each turn, we plot the mean of the empirical strategy distribution. Probing on odd persuader turns has a relatively identical strategy curve compared to GPT-4.1-Nano, the reference curve we're comparing against. All plots show logical appeal as not dominant across all samples. The probe has competing ER and EE oscillations which smoothens out if we only consider even/odd turns. Note that as conversation length increases beyond 23 turns (a small subset of the total dataset), emotional appeal is the dominant strategy which is consistent across all models. A single turn corresponds to a single EE/ER utterance. In Figure~\ref{fig:jsd-strategy-dist}b, we take a point-wise Jensen-Shannon Distance at each turn using the mean probability vectors in this plot.}
    \label{fig:strategy}
\end{figure*}

\begin{figure*}[htb]
    \includegraphics[width=1.00\textwidth]{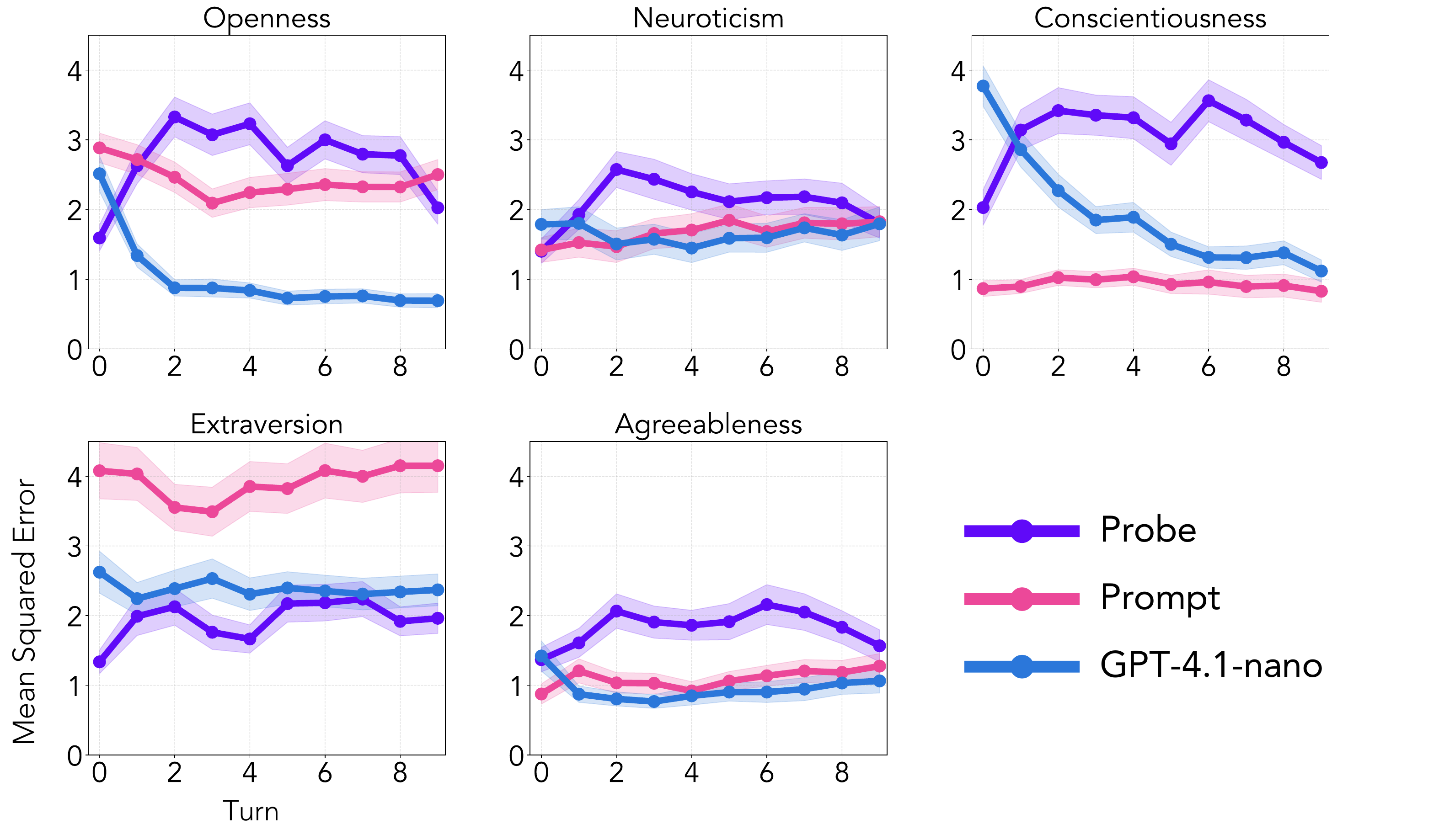}
    \caption{\textbf{{Persuadee Personality Predictions}}}
    \label{fig:ee-personality}
\end{figure*}

\begin{figure*}[htb]
    \includegraphics[width=1.00\textwidth]{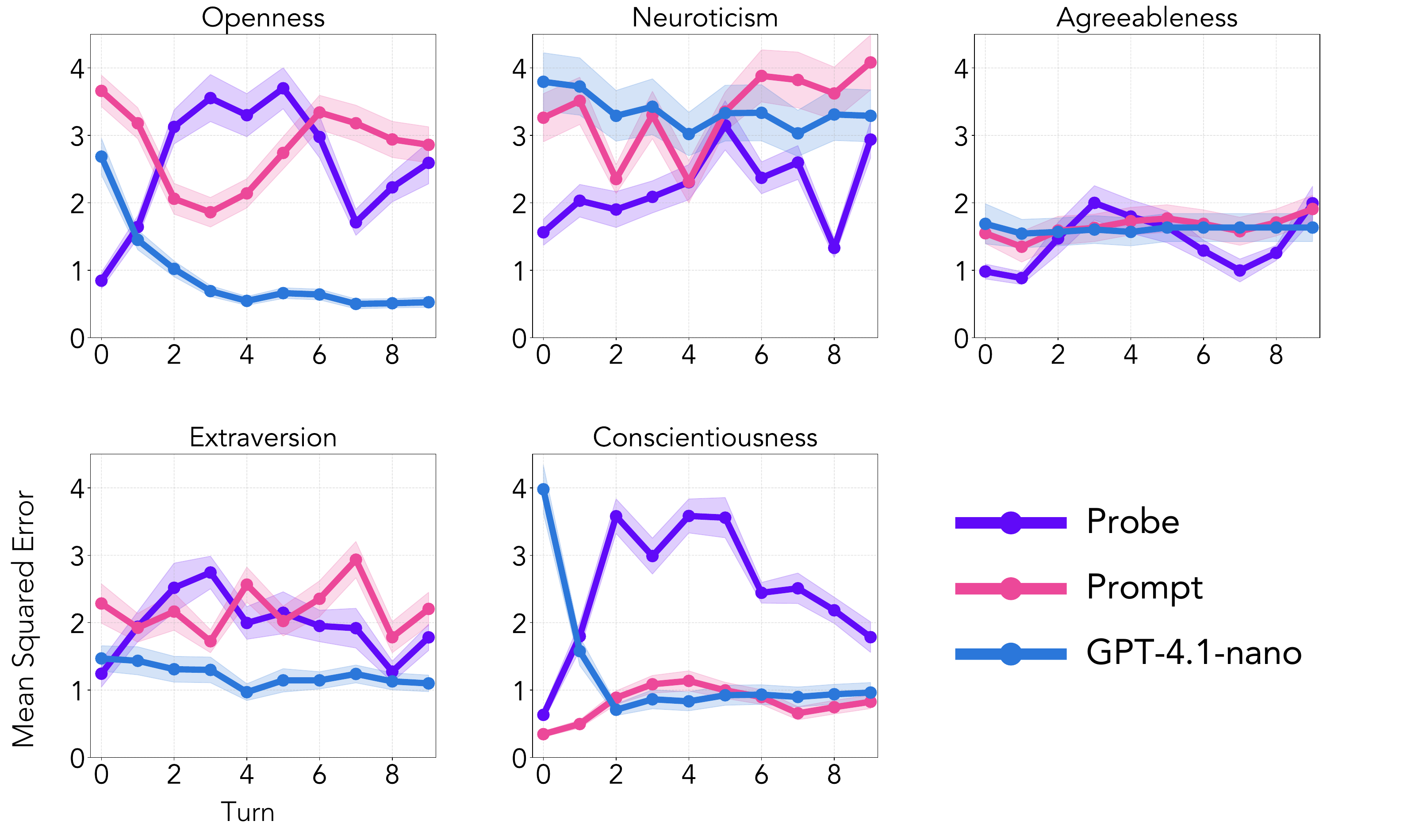}
    \caption{\textbf{{Persuader Personality Predictions}}}
    \label{fig:er-personality}
\end{figure*}

\begin{figure}[htbp]
  \centering
  \includegraphics[width=1.0\columnwidth]{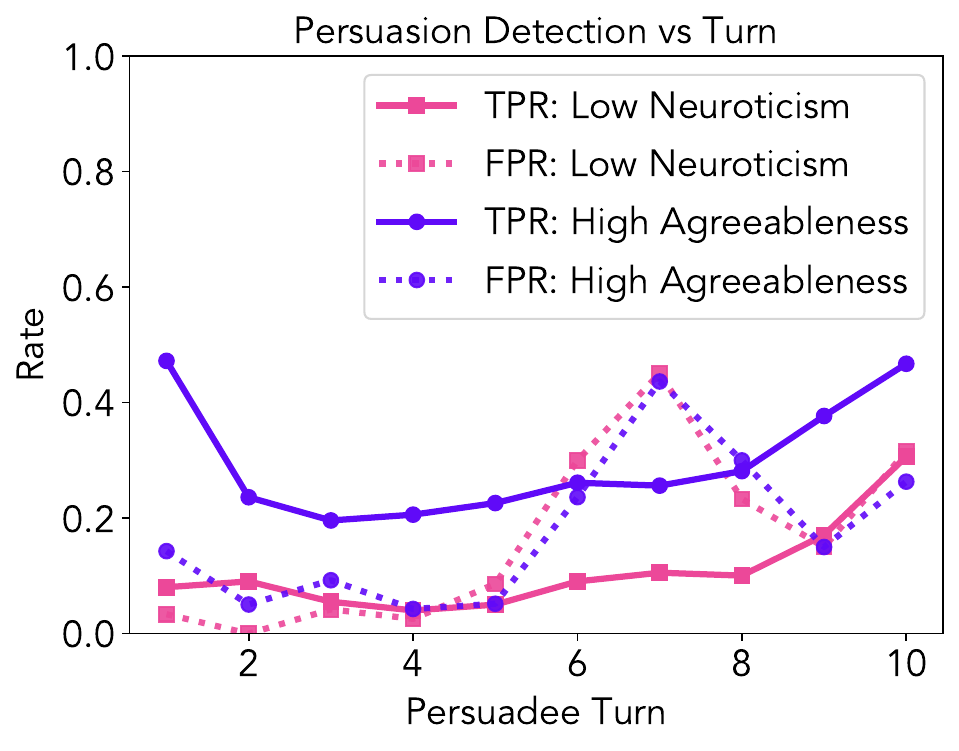}
  \caption{\textbf{Persuasion detection vs turn} represents the inverse of Figure~\ref{fig:detection-perf-forward}: we test whether low neuroticism (thresh < 0.2) and high agreeableness (thresh > 0.8) can accurately identify persuasive dialogues (TPR) while minimizing unpersuaded false alarms (FPR). Neuroticism fails—the FPR curve consistently exceeds TPR, misflagging more unpersuaded than persuaded samples. Agreeableness does better with modest TPR peaks at the first and last turns, though not great overall.}
  \label{fig:detection-perf-backward}
\end{figure}

\begin{itemize}
    \item Figure~\ref{fig:strategy} is the extension to Figure~\ref{fig:jsd-strategy-dist} which shows the full strategy trajectory curves for each model. Figure~\ref{fig:jsd-strategy-dist} is computed from this by taking a point-wise JSD between probe/prompt model and GPT-4.1-Nano, for a given turn.
    
    \item Figures~\ref{fig:ee-personality} and ~\ref{fig:er-personality} are full personality trajectory plots for each trait, extending Figure~\ref{fig:ee-personality-main}. 
    
    \item Figure~\ref{fig:detection-perf-backward} extends Figure~\ref{fig:detection-perf-forward} and provides the negative result of the inverse problem discussed in Section~\ref{sec:detect-pers-personality}.
\end{itemize}

\newpage
\section{Sample Sizes and Statistical Significance}
\label{app:stat-sig}

Table~\ref{tab:fig-stat-sig} outlines sample sizes and statistical significance for figures and experiments used in this paper.

\begin{table}[htbp]
  \centering
  \caption{Samples Sizes used in Figures}
  \label{tab:fig-stat-sig}
  \begin{tabularx}{\linewidth}{@{} l X @{}}
    \toprule
    \textbf{Figure} & \textbf{Sample Size} \\
    \midrule
    Figure~\ref{sample-traj}
      & Random samples. \\
    Figure~\ref{fig:auroc}
      & PfG: $n=401$ (232 persuasion, 169 unpersuasion); DP: $n=186$ (93 per class, due to downsampling). \\
    Figure~\ref{fig:timing}
      & Timing comparison: synthetic token‐level ($n=5$), turn‐level ($n=10$), conversation‐level ($n=20$). Error bars represent average across 3 runs. \\
    Figure~\ref{fig:interaction}
      & DP: $n=557$, PfG: $n=545$ (all positive donation samples) \\
    Figures~\ref{fig:strategy},~\ref{fig:jsd-strategy-dist}
      & PfG: $n=200$ for Probe/Llama-3b, $n=140$ for GPT-4.1-Nano. Error bands represent \(0.3\,\sigma\). \\
    Figures~\ref{fig:ee-personality-main}, \ref{fig:ee-personality}
      & PfG: $n=100$. Error bands represent \(0.1\,\sigma\). \\
    Figure~\ref{fig:er-personality}
      & PfG: $n=20$ for Probe/Prompt, $n=40$ for GPT. Error bands represent \(0.1\,\sigma\). \\
    Figures~\ref{fig:detection-perf-forward}, \ref{fig:detection-perf-backward}
      & PfG: $n=401$ (since we filter to use annotated subset)\\ 
    \bottomrule
  \end{tabularx}
\end{table}

\subsection{Upsampling to increase the amount of unpersuasive samples}
\label{app:prompt-templates-unpersuasive}

As mentioned in Section~\ref{sec:datasets}, our evaluation datasets consist of predominantly positive samples. To remedy this, we few-shot prompt GPT-4.1 to increase the amount of unpersuasive samples. Table~\ref{tab:prompt-templates-unpersuasive} shows the prompt template we use to upsample PfG unpersuasive samples (we do something similar for DP).

\begin{table}[ht]
  \centering
  \caption{Prompt Template for Unpersuasive Donation Conversations}
  \label{tab:prompt-templates-unpersuasive}
  \begin{tabularx}{\columnwidth}{@{}l X@{}}
    \textbf{System:}
      & You are a helpful assistant that writes realistic, multi-turn conversations
        in which a persuader fails to convince the persuadee to donate. \\[0.75em]
    \textbf{User:}
      & Below are \emph{N} examples of conversations where the persuader fails to
        persuade the persuadee to donate (unpersuasive conversations): \\[0.5em]
      & \quad Example 1: \texttt{<Unpersuasive example 1 (omitted for brevity)>} \\
      & \quad Example 2: \texttt{<Unpersuasive example 2 (omitted for brevity)>} \\
      & \quad \dots \\[0.5em]
      & Please generate \emph{k} independent conversations in the same style.
        Each should be multi-turn, realistic, and unsuccessful at persuading
        the persuadee to donate.
  \end{tabularx}
\end{table}

\newpage
\section{Probe Learning Curves}
\label{app:learning_curves}

\begin{figure*}[h]
  \centering
  \includegraphics[width=0.70\textwidth]{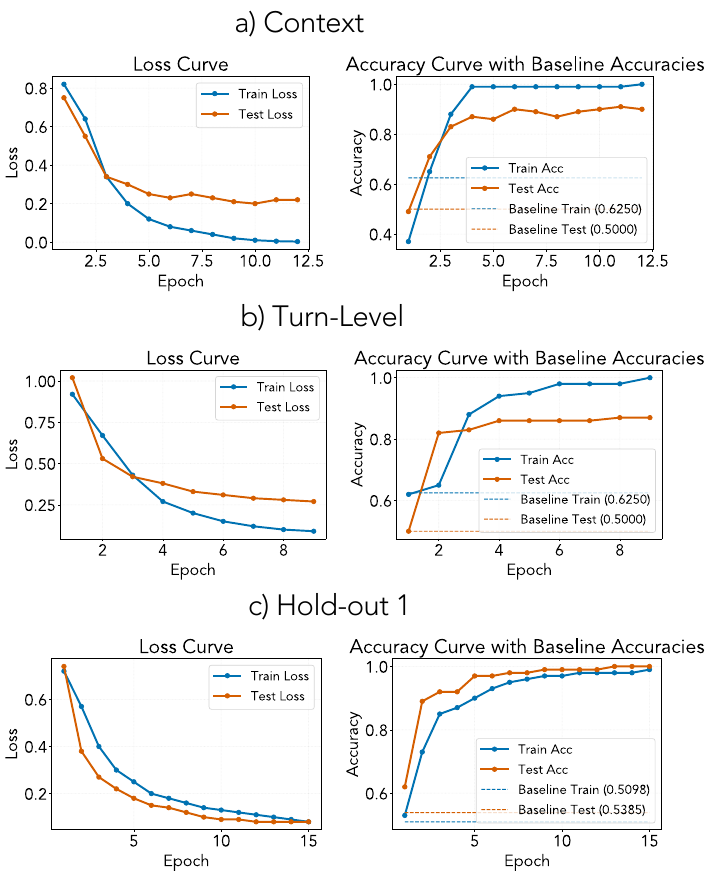}
  \caption{\textbf{Probe training curves across various settings.} The context probe a) is what we use in the main body of the paper. Additionally, we train (b) turn-level (contextless) probes which are trained on individual turns rather than entire conversations and (c) hold-out-1 probes which are trained on the entire conversation with the exception of the last turn. The idea behind (c) was that we observed most of the persuasion signal coming from the final conversation turn so removing this final turn could help mitigate a potential spurious correlation. However, probes a) and b) did not latch onto this spurious correlation and thus c) ended up under-performing compared to them.}
  \label{fig:probe-training-curves}
\end{figure*}

In Figure~\ref{fig:probe-training-curves} we outline and show learning curves for various other linear probe configurations, aside from the context-based probe we use in the main paper. 

These probes were all trained using the Adam optimizer with a learning rate of $1\times10^{-3}$, ReLU activation function, and the standard cross-entropy loss.

\newpage

\section{Data Analysis and Validation}

\subsection{Validating where persuasion occurs in a conversation}

\begin{figure*}[h]
  \centering
  \includegraphics[width=0.80\textwidth]{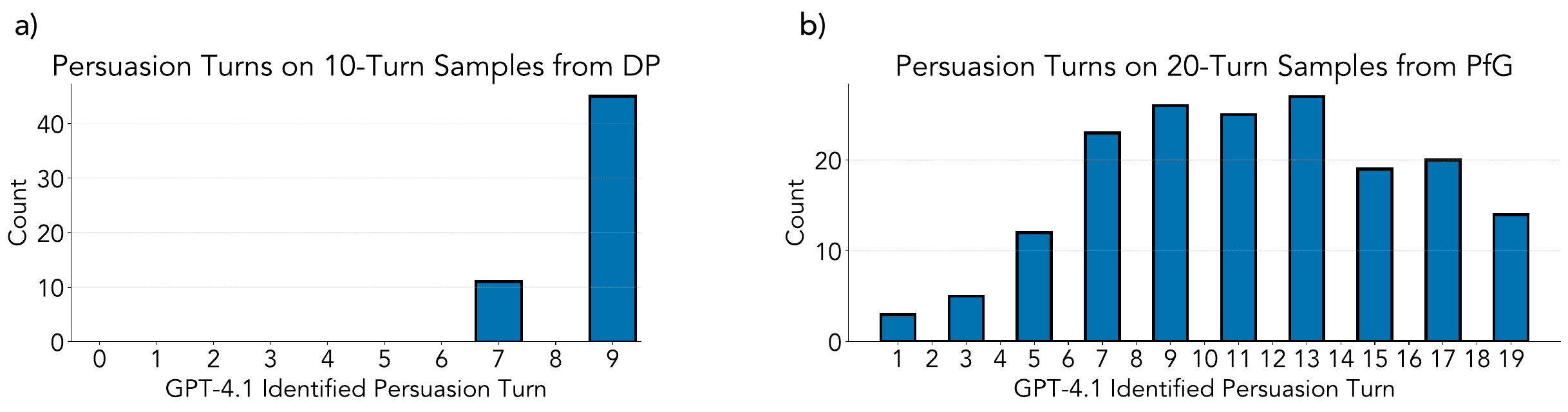}
  \caption{\textbf{Persuasion Turn Distributions} for \textbf{a)} DailyPersuasion (DP) and \textbf{b)} PersuasionforGood (PfG). These histograms show the frequency of successful persuasion moments by conversation turn. In DP, persuasion predominantly occurs in the final turn, while PfG exhibits a more even dispersion across the middle turns.}
  \label{fig:pers-turn-dist}
\end{figure*}

At several places in our paper, such as in Figure~\ref{fig:auroc} we claim that in DailyPersuasion (DP), persuasion outcomes typically occur in the final 1-2 persuadee turns of a conversation, while in PersuasionforGood (PfG), they are more evenly dispersed throughout the middle-to-late turns. Here, we systematically test these claims by zero-shot prompting GPT-4.1 using the prompt template in Figure~\ref{fig:prompt-template-turn-classification} to classify the turn where persuasion occurs. We use 56 random samples from DP and 100 from PfG. The first-author manually validated roughly 25 classifications with no disagreements on DP and 4/26 disagreements on PfG, indicating an overall high agreement. The resulting persuasion turn histograms are shown in Figure~\ref{fig:pers-turn-dist} which corroborate our claims.

\begin{figure}[ht]
  \centering
  \begin{verbatim}
System:
Analyze when the persuadee was persuaded in this conversation. 
Look for the specific turn where the persuadee’s attitude shifted or they became convinced.

User:
Turn 0 (Role): <first utterance>
Turn 1 (Role): <second utterance>
…
Turn N (Role): <Nth utterance>

Which turn number shows the persuadee being persuaded?
Reply with just the turn number (e.g. “3”) or “None” if no persuasion occurred.
  \end{verbatim}
  \caption{System and user messages used to prompt GPT-4.1 for persuasion turn identification.}
  \label{fig:prompt-template-turn-classification}
\end{figure}

\subsection{DailyPersuasion (DP) Data Analysis}
\label{dp-distr-stats}
Here we show distributional statistics and an annotated conversation for DP \cite{jin-etal-2024-persuading}.

\begin{figure*}[htb]
  \centering
  \includegraphics[width=0.8\textwidth]{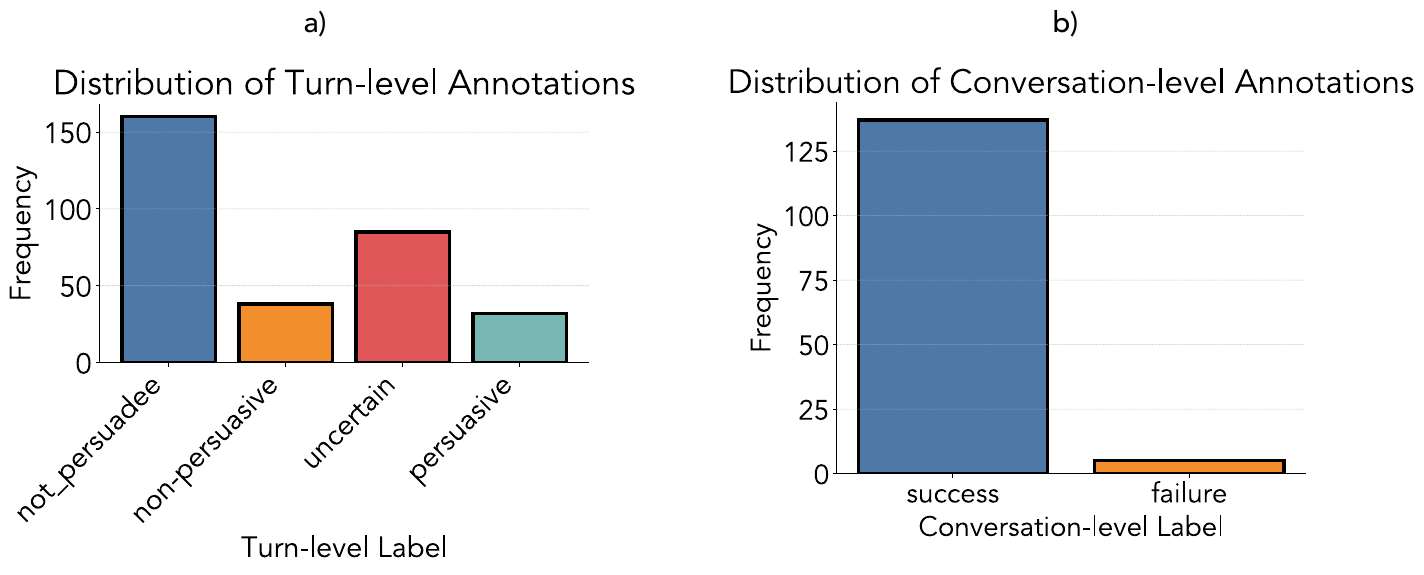}
  \caption{\textbf{Using GPT-4o to annotate DailyPersuasion turns and conversations.} Most conversations have positive persuasion motivating our choice to use GPT to up-sample failure conversations in Section~\ref{sec:datasets}. The persuasive/non-persuasive samples in turn-level annotations are used for the contextless evaluations in Table~\ref{tab:daily_persuasion_results} (\S \ref{app:individual-utterance-eval}).}
  \label{fig:dp-label-distr}
\end{figure*}

Figure~\ref{fig:dp-label-distr} shows distributions over both turn-level and conversation-level annotations generated by GPT-4o. We manually review these and find high human annotator agreement (Cohen's $\kappa > 0.95$). See Section~\ref{app:dp-example} for a sample annotation.

\subsubsection*{Fully Annotated Success Conversation from DP}
\label{app:dp-example}
\begin{Verbatim}[fontsize=\small]
Fully Annotated Conversation, Conversation ID: 20230831164028-67_95_2
Session-Level Annotation (gpt-4o): success
Turn-Level Annotations (only for Persuadee turns):
    Turn 1: not_persuadee
    Turn 2: non-persuasive
    Turn 3: not_persuadee
    Turn 4: uncertain
    Turn 5: not_persuadee
    Turn 6: non-persuasive
    Turn 7: not_persuadee
    Turn 8: uncertain
    Turn 9: not_persuadee
    Turn 10: uncertain
    Turn 11: not_persuadee
    Turn 12: persuasive
Dialogue Session:
    Persuader: Hey Sam, did you know that volunteering at the local food bank can 
        actually improve your problem-solving and teamwork skills? It's just like 
        leveling up in a video game!
    Persuadee: Really? But I still think I'd rather spend my weekends playing 
        games. I don't see how volunteering can be that beneficial.
    Persuader: You know, our friend Jessie started volunteering at the food bank 
        last year. She told me that it helped her develop leadership skills, and 
        she even got a scholarship for her community involvement! It's worth 
        giving it a try, don't you think?
    Persuadee: Hmm, I didn't know that. But what if I don't enjoy it? I'm not 
        really a people person, and it sounds like a lot of work.
    Persuader: I totally get that, but the great thing about volunteering at the 
        food bank is that we'll be working together as a team. Plus, you'll meet 
        other people who share your interests, and it's always more fun to do 
        things with friends, right?
    Persuadee: Yeah, maybe. But I still think I'd rather just play video games. 
        I don't feel like I have much to offer.
    Persuader: Sam, you have so much to offer! Your gaming skills show that 
        you're quick-thinking and resourceful, which are valuable qualities in a 
        volunteer. Plus, you'll be making a real difference in people's lives by 
        helping out at the food bank.
    Persuadee: Well, maybe you're right. But I still have concerns about the time 
        commitment. I don't want it to interfere with my gaming time.
    Persuader: No worries, Sam! The food bank has flexible volunteering schedules, 
        so you can choose a time that works best for you. You can still enjoy your 
        gaming time and make a difference in the community.
    Persuadee: Okay, that sounds more doable. Maybe I can give it a try, but I'm 
        not making any promises.
    Persuader: That's totally fine, Sam! Why not give it a try for just one or two 
        weekends? If you don't like it, you can always stop, but at least you'll 
        have given it a shot. I'll be there with you, and we can make it fun 
        together!
    Persuadee: Alright, I'll give it a try. But only for a couple of weekends, 
        and then we'll see how it goes.
\end{Verbatim}

\newpage
\section{Evaluating Probes on Individual Utterances}
\label{app:individual-utterance-eval}

In this section, we evaluate how accurately various probe configurations (outlined in Figure~\ref{fig:probe-training-curves}, \S \ref{app:learning_curves}) can categorize individual persuadee utterances as ``persuaded" or ``not-persuaded" without access to the broader conversation context. We focus on the DailyPersuasion (DP) dataset, using the label distribution shown in Figure~\ref{fig:dp-label-distr}. For comparison, we also assess performance on (1) the full conversation, and (2) the full conversation with the final turn removed—since, as Figure~\ref{fig:auroc} indicates, much of the persuasion signal in DP is concentrated in that last turn, potentially creating a spurious correlation to classifying persuasion.

\begin{table*}[ht]
    \centering
    \caption{Comparison of probe performance on the DailyPersuasion dataset. Ultimately, the context probe does best, motivating our use of it in the main experiments of this paper.}
    \resizebox{\textwidth}{!}{
        \begin{tabular}{l c c c c c c c c c}
        \toprule
        \multirow{2}{*}{\textbf{Probe Configuration}} 
            & \multicolumn{3}{c}{\textbf{Turn Classification}} 
            & \multicolumn{6}{c}{\textbf{Conversation Classification}} \\
        \cmidrule(lr){2-4} \cmidrule(lr){5-10}
         & Accuracy & Precision & Recall 
            & \multicolumn{3}{c}{Full conversation} 
            & \multicolumn{3}{c}{Last turn omitted} \\
        \cmidrule(lr){5-7} \cmidrule(lr){8-10}
         &  &  &  & Accuracy & Precision & Recall 
            & Accuracy & Precision & Recall \\
        \midrule
        \textbf{Context*}    & 91 & 91 & 91 & 95 & 96 & 99 & 95 & 96 & 99 \\
        No-context & 68 & 60 & 97 & 97 & 97 & 100 & 89 & 97 & 92 \\
        Hold-1     & 83 & 86 & 75 & 96 & 96 & 99 & 95 & 96 & 99 \\
        Hold-2     & 71 & 80 & 50 & 94 & 97 & 97 & 76 & 98 & 77 \\
        \bottomrule
        \end{tabular}
    }
    \label{tab:daily_persuasion_results}
\end{table*}

\textbf{Results:} Table~\ref{tab:daily_persuasion_results} shows the result of our probe evaluations on DP. Ultimately, the context probe does well across all scenarios, motivating our use for it in the main experiments of this paper. The no-context probe is heavily biased towards predicting positive persuasion for most turn-level samples (low precision, high recall). It does well on conversation classification since most samples are positive persuasion. 

\newpage
\section{Persuasion Probe Interpretability}

In this section, we more deeply investigate what our persuasion probe learns about persuasion (Figures~\ref{fig:semantic-labels}, \ref{app:token-ablations}) as well as the effect of probing various layers on persuasion classification performance (Figure~\ref{app:acc-vs-layer}).

\subsection{Experiments}
\label{app:interp-exps}
\begin{figure*}[htb]
    \includegraphics[width=1.00\textwidth]{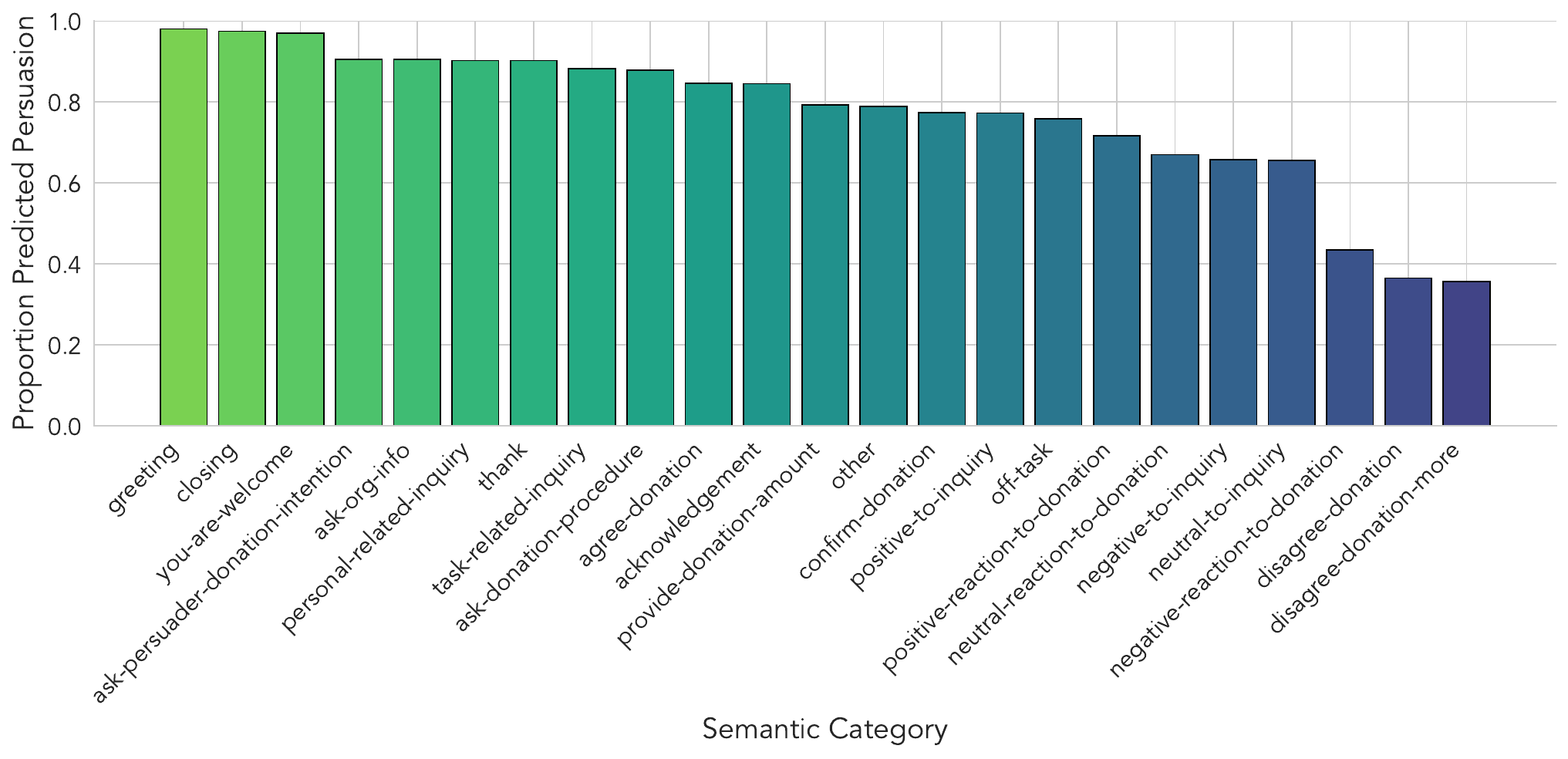}
    \caption{\textbf{We apply our persuasion probe to classify individual EE utterances as persuasive/unpersuasive.} The y-axis shows the proportion of samples in each bin that were classified as persuasive (0.5 threshold). Categories with a higher proportion of persuasive classifications are towards the left. This shows that \textbf{our probe is well-calibrated} since persuasive categories such as ``agree-donation" are towards the left, unpersuasive categories such as ``disagree donation" are towards the right and neutral categories like ``other" are towards the middle.}
    \label{fig:semantic-labels}
\end{figure*}

\textbf{Utterance-Level Classification:} How well does our persuasion probe do at picking out individual persuasive/unpersuasive turns from PfG? In PfG, each EE utterance has a semantic label. We can use these labels to better understand what the probe labels as (un)-persuasive text. Figure~\ref{fig:semantic-labels} shows a histogram for our persuasion probe where the Y-axis shows the proportion of samples in each bin that were classified as persuasive (0.5 threshold). Categories with a higher proportion of persuasive samples are towards the left. We expect semantically negative categories like disagree-donation, negative-to-inquiry, and negative-reaction-to-donation to correspond with a lower p(persuasion) and vice-versa for positive categories.

In Figure~\ref{fig:semantic-labels} positive labels are towards the left and negative labels are towards the right as expected. We still see some negative labels like disagree-donation get classified as persuasion and some positive labels classified as unpersuasion. We speculate this may be due to some samples having short lengths, indicating not enough signal to determine persuasion. It may also be due to incorrect human annotations. For instance, in the negative-reaction-to-donation bin we see samples like: ``I prefer to volunteer if I can" and ``Amounts don't matter to me.". The latter actually indicates that there can be persuasive instances in seemingly negative label categories. We see ``other" and ``neutral-to-inquiry" near the middle as expected (i.e. a neutral category should have near 50\% accuracy).

\begin{figure*}[htb]
    \centering
    \includegraphics[width=1.0\textwidth]{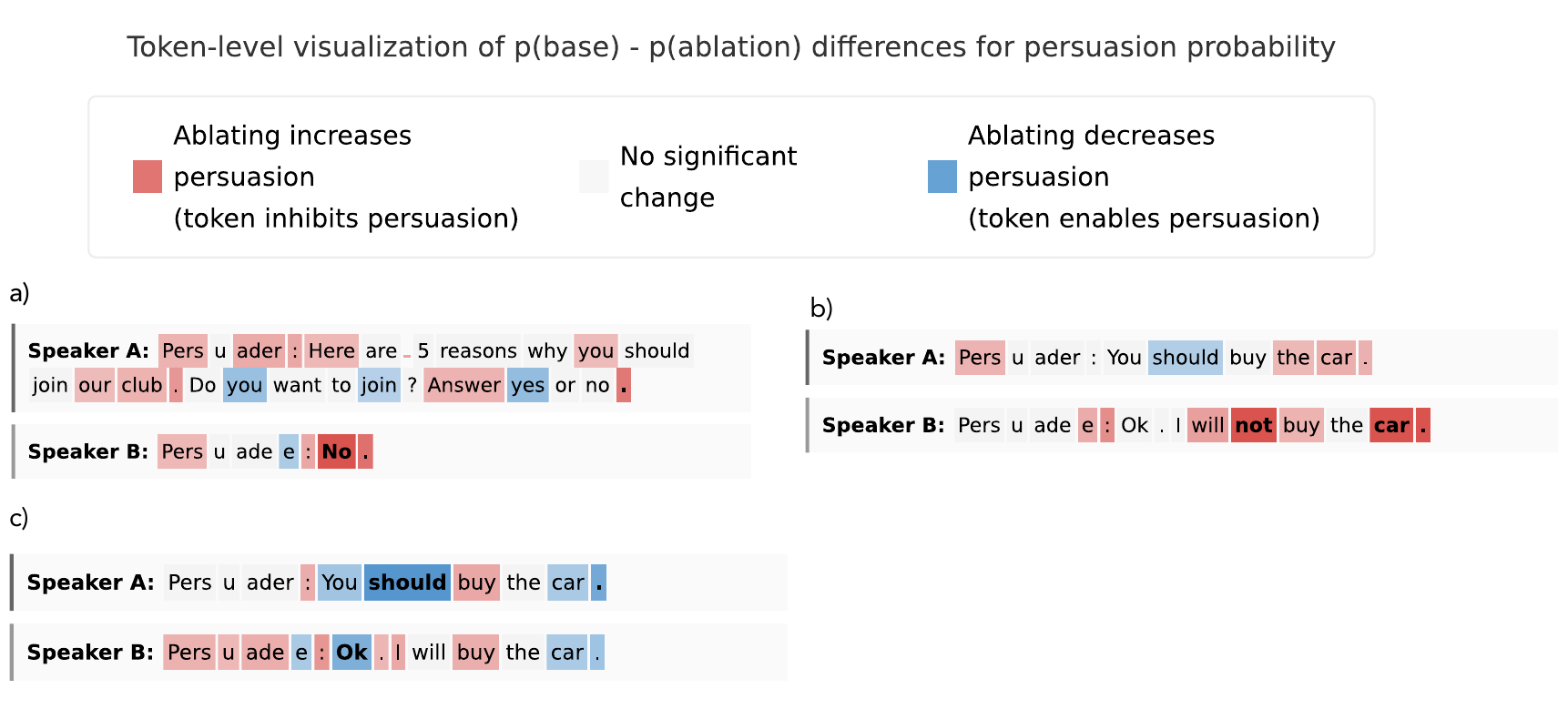}
    \caption{\textbf{Knock-one-out token ablations.} We measure the effect on $p(\text{persuasion})$ when ablating a specific token. Red denotes an increase in $p(\text{persuasion})$, blue a decrease. As expected, in a) and b), ablating negative tokens like No/Not produce strong effects as they change the persuasive content of the message. Ablating positive tokens ``Ok" and ``will" in c) has a weaker effect suggesting the model can still infer persuasive intent even when one of those tokens are removed.}
    \label{app:token-ablations}
\end{figure*}

\textbf{Token attribution:} In Figure~\ref{app:token-ablations}, we analyze how ablating individual tokens affects $p(\text{persuasion})$ across both unpersuasive (\ref{app:token-ablations}a, \ref{app:token-ablations}b) and persuasive samples (\ref{app:token-ablations}c). As expected, ablating negative tokens like ``No" and ``Not" produce strong effects since they change the persuasive content of the message. In \ref{app:token-ablations}c ablating positive tokens like ``Ok" and ``buy" has a weaker effect perhaps suggesting the model can still infer persuasive intent even when one of those tokens are removed. For instance, in the unpersuasive cases, deletion of No/Not flips semantic meaning from unpersuaded to persuaded whereas no single deletion in the persuasive case achieves this reversal.

\begin{figure*}[htb]
    \includegraphics[width=1.00\textwidth]{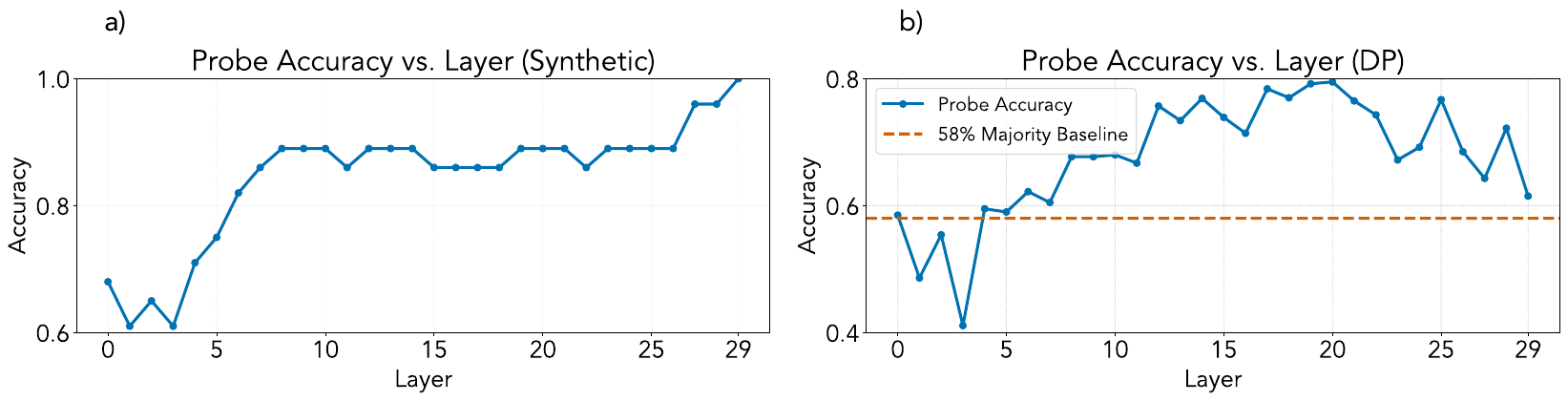}
    \caption{\textbf{Probe Layer vs. Accuracy in a) our synthetic training environment vs. (b) DailyPersuasion.} Probing the middle-to-late layers yields the best performance on DailyPersuasion. Motivated by this, we extract activations from layer 26/30 in our main paper.}
    \label{app:acc-vs-layer}
\end{figure*}

\textbf{Effect of probing layer on performance:} In Figure~\ref{app:acc-vs-layer}, we investigate the effect of the probing layer on persuasion predictive performance. Probing the middle-to-late layers generalizes best. Motivated by this, we extract activations from layer 26/30 in our paper.

\newpage
\section{Conversation Transcripts}
\label{app:transcripts}

Here we provide conversation transcripts to supplement descriptions and figures in the main paper.

\subsection{Persuasion Probability Samples for Figure~\ref{sample-traj} a)}
\label{app:persuasion-transcript}
\subsubsection*{Successful Persuasion}
\begin{persuasiondialogue}
    \persuader{Hey, how's it going tonight?}
    \persuadee{Very Well, How about you?}
    
    \persuader{Ok, here too. I'd like to talk to you about children who are less fortunate. Do you have a moment?}
    \persuadee{Yes, please go ahead and tell me}
    
    \persuader{How do you feel about people who just born unlucky?}
    \persuadee{I feel sorry for them and would like to help them as much as possible}
    
    \persuader{I see I got a really kind person to talk to. That's great.}
    \persuadee{Go ahead and tell me how can I help}
    
    \persuader{I'm glad you're so welling to help, but I would feel bad just to get your help without letting you know what the organization is about.}
    \persuadee{Sure, please tell me about the organization}
    
    \persuader{It's an organization called Save the Children. Have you heard of them?}
    \persuadee{I think I have}
    
    \persuader{They are an international non-governmental organization that promotes children's rights, provides relief and helps support children in developing countries.}
    \persuadee{that is very good to know}
    
    \persuader{Yes. Just recently, not sure if you've heard of it, they are working on reuniting children who were separated from their parents at the Mexico-US border.}
    \persuadee{Thats such a noble thing they are doing}
    
    \persuader{They are very noble. They're helping kids worldwise but also very close to home. Would you like to help by donating?}
    \persuadee{Sure, I would like to help. How much is the average donation}
    
    \persuader{They get anything from \$0.50 to \$2 from mturk workers.}
    \persuadee{I would like to to donate \$1.00 as a mturk worker}
\end{persuasiondialogue}

\subsubsection*{Unsuccessful Persuasion}

\begin{persuasiondialogue}
    \persuader{Hello there—how are you?}
    \persuadee{I am good. How are you?}

    \persuader{I'm pretty good, just trying to get ready for work soon.}
    \persuadee{I understand that. So tell me about this charity.}

    \persuader{Have you heard of Save the Children? They do work around health, education, etc.}
    \persuadee{I'm not sure that I have.}

    \persuader{They work mostly in developing countries. I would appreciate it if some of the funds could be donated. The researchers will pass that donation on to the charity at the end.}
    \persuadee{What kind of work do they do?}

    \persuader{They do emergency hunger relief, are engaged with Syrian children.}
    \persuadee{Ok, that sounds good. How do I know it is a legitimate charity?}

    \persuader{You can check their website URL. They are highly rated by Charity Watch.}
    \persuadee{That's good to know. How long have they been around?}

    \persuader{Almost 100 years!}
    \persuadee{That's pretty impressive.}

    \persuader{Do you think you would be willing to donate a portion of your earnings?}
    \persuadee{I just can't see doing that.}

    \persuader{Why do you think that? There are children dying in Syria who can benefit from the donation.}
    \persuadee{I understand that. But, I choose my own charities independently.}

    \persuader{Ok, I understand that. I hope you have a good day.}
    \persuadee{Ok, you too!}
\end{persuasiondialogue}

\subsection{Credibility Appeal Sample for Figure~\ref{sample-traj} b)}
\label{app:credibility-transcript}
\begin{persuasiondialogue}
    \persuader{Hello, how are you today?}
    \persuadee{Hi, it was another day}

    \persuader{I see, same here for me. Have you ever tried donation through HITs?}
    \persuadee{Ya, recently I too tried children's charity regarding their studies. It makes me feel good}

    \persuader{I usually donate a good part of my bonus in different studies, but I especially have a soft heart for children. Have you ever heard about “Save the Children” before?}
    \persuadee{Children must be educated. I didn't hear about “Save the Children.” What is the procedure for giving a donation?}

    \persuader{They work on different issues in almost 120 countries. They focus on education, hunger, health, and emergency relief.}
    \persuadee{Really? That’s pleasing to hear. Who’s taking care of this charity, and how do they manage in 120 countries?}

    \persuader{They either host events or travel to those countries. There are big partners like IKEA and Walt Disney Studios. You can read more here: URL/s/about-us/leadership-and-trustees}
    \persuadee{Sure, I will check this link. Is there a procedure to join as a member in that group? Can a normal person from the middle class join?}

    \persuader{I believe everyone can join to help. I’ve talked to someone who worked there and she already traveled to Africa. I’d love to try but don’t have enough days off.}
    \persuadee{People with kind hearts must join as members or sponsor a child. I’ve referred some friends in my circle.}

    \persuader{I know, right? It’s a big commitment—some people dedicate their whole lives. I like how through HITs you can even help with just one cent.}
    \persuadee{I asked earlier: what is the procedure to donate?}

    \persuader{On the website there is a link to donate: \url{https://support.URL/Donation2?df_id=1620&1620.donation=form1}}
    \persuadee{Okay, I see. Is there any disclosure of our identity when donating, large or small?}

    \persuader{No, it will be anonymous. You can also donate at the end of this HIT. It can be any amount—even \$0; you’ll be asked.}
    \persuadee{Whether I can trust this charity? From the beginning you talked only about this charity. Are you a member? Are there no other children’s charities?}

    \persuader{I’ve donated before—they have clear financial reports and reputable partners. I trust them. I’m not a member, just a donor. There are a few other charities too.}
    \persuadee{Oh, that’s nice of you. I’ll form my own opinion and donate as much as I can, and share with my friends.}

    \persuader{Thank you for your time and effort. It was a pleasure chatting with you. Have a good night, take care!}
\end{persuasiondialogue}

\subsection{Personality Samples for Figure ~\ref{sample-traj} c) and d)}
\label{app:personality}
\subsubsection*{Successful Persuasion}

\begin{persuasiondialogue}
    \persuader{Hello...How are you this morning?}
    \persuadee{I'm doing very well thanks, got these kids off to school this morning without a hitch. How are you?}

    \persuader{I'm great...Speaking of children, are you familiar with the organization Save The Children?}
    \persuadee{I think I've heard of them; are you a member or on the board?}

    \persuader{I am a member of the organization. Save The Children is an international non-governmental organization that promotes children's rights, provides relief, and helps support children in developing countries.}
    \persuadee{Would you be able to provide me some information to have a further look?}

    \persuader{You can refer to Save The Children's website to gather more information if you like: \url{URL}}
    \persuadee{Thank you so much for that. Are you an MTurk worker or an administrator of the HIT?}

    \persuader{Administrator of the HIT. Today we are looking for people interested in donating some of, or all of, their HIT reward to Save The Children.}
    \persuadee{Well, it's a great cause so I'd be willing to donate \$0.50 to Save The Children.}

    \persuader{How much would you like to donate to the charity now? Your donation will be directly deducted from your task payment. You can choose any amount from \$0 to all your payment (\$2).}
    \persuadee{I can donate \$0.50. I'm a caregiver for a child with different abilities, so I hope that helps.}

    \persuader{Every bit helps...Thank you so much!}
    \persuadee{You're welcome. Do the majority choose to donate or do most keep their full payment? I'm just curious.}

    \persuader{A lot of people choose to donate.}
    \persuadee{Well, I'm glad to hear that. On another note, as soon as I'm finished with HITs I think I'll take a look at the link you gave me and see what else I can do. How long have you been administering HITs?}

    \persuader{I've been administering HITs for a few months so far. Thank you for taking the time to review the website and see what else you can do!}
    \persuadee{Thank you very much and good luck with your cause. Have a great day. :-)}

    \persuader{Thanks and I hope you have a wonderful Friday!}
    \persuadee{Same to you, thanks for the opportunity.}
\end{persuasiondialogue}

\subsubsection*{Unsuccessful Persuasion}
\begin{persuasiondialogue}
    \persuader{hello how are you?}
    \persuadee{I'm good. How is your day going?}

    \persuader{good. Would you like to donate to help save the children?}
    \persuadee{I don't think I am currently able to give anything.}

    \persuader{what do you mean? why not?}
    \persuadee{Things are so tight at home that I can barely make things work as it is.}

    \persuader{seriously? You can't even donate 0–2 dollars for other kids?}
    \persuadee{Not if it takes away from my own kids that are my responsibility.}

    \persuader{what about the kids that really need it though? why don't you think about them?}
    \persuadee{Well I am thinking about them. I don't really know what kids this is even going to though.}

    \persuader{but people donate all the time without knowing who it is going to}
    \persuadee{Well that seems silly. You wouldn't know if your money is going to a child or some random people's pocket.}

    \persuader{can you please donate? one of the children in this group is my child}
    \persuadee{Well that doesn't really legitimize the operation.}

    \persuader{how about just a dollar?}
    \persuadee{I would need to know what organization it was going to.}

    \persuader{here: Save the Children is an international non-governmental organization that promotes children's rights, provides relief and helps support children in developing countries.}
    \persuadee{What percentage of the donation actually goes towards helping the children?}

    \persuader{one hundred percent; why would someone take that money for themselves?}
    \persuadee{Because all organizations take a percentage to pay for overhead costs.}

    \persuader{how about just 75 cents then?}
    \persuadee{No I won't be contributing today.}
\end{persuasiondialogue}

\subsection{Conversation Samples from PersuasionforGood}
\label{app:pfg-samples}
\subsubsection*{a)}
\begin{persuasiondialogue}
    \persuader{We are collecting money for an important children's charity.}
    \persuadee{it is great things to do}

    \persuader{We would like you to help.}
    \persuadee{sure, but first i need to know for what need will i donate and what organization}

    \persuader{It is for the charity Save the Children. You are probably familiar with it.}
    \persuadee{yes, i have heard. How can i donate? in what way?}

    \persuader{The researchers are looking for people to donate part of their payment for this study to this charity. Can you contribute?}
    \persuadee{do you mean autopay from my payment checks ?}

    \persuader{No, I think they mean reducing your payment for this study and giving the difference to Save the Children.}
    \persuadee{okey. i got it. i am interesting}

    \persuader{Thank you, it will help out children in need.}
    \persuadee{yes. what do i need to do now to help?}

    \persuader{I suspect that in the following part of the study, the researchers will ask how much of the money you want to donate.}
    \persuadee{okey. i understad you}

    \persuader{How much money are you going to donate? I have to admit these mturk studies don't even pay you enough for a cup of coffee.}
    \persuadee{i will donate now 5 dl}

    \persuader{Wow, that's much more than the study is supposed to pay me! Thank you.}
    \persuadee{i am alway happy to help}

    \persuader{I believe this completes my part of the study. Thank you again, and have a great day.}
    \persuadee{have a great day too}
\end{persuasiondialogue}

\subsubsection*{b)}

\begin{persuasiondialogue}
    \persuader{Hello?}
    \persuadee{hi}

    \persuader{How are you today?}
    \persuadee{good how about you?}

    \persuader{I am well, thank you! Glad you are doing well also.
               Have you heard of the charity called Save the Children?}
    \persuadee{i have not, can you tell me more?}

    \persuader{It is an international charity that helps children in
               need all around the world. These children have been
               victims of war, famine, and poverty. Does this sound
               like a good charity?}
    \persuadee{definitely, have you donated to them?}

    \persuader{I have. 86\% of the donations go directly to the
               programs that help the children. Do you believe that
               children in need deserve help from others?}
    \persuadee{yes i do}

    \persuader{Great! Donations made from people like you can directly
               improve the lives of thousands of children all over the
               world. Even a small donation makes a big difference.
               Have you ever donated money to a charity before?}
    \persuadee{yes i have. a few times}

    \persuader{That's great! Donating your payment from this task to
               Save the Children would make a big difference in a
               child's life today. You said earlier you believe
               children in need deserve help from others. Would you be
               willing to donate your payment from this task to the
               charity to help a child?}
    \persuadee{sure would love to}

    \persuader{Awesome! The research team collects all donations from
               this task and send them to Save the Children. How much
               of your payment would you be willing to donate? You can
               choose any amount from \$0 to the entire payment.}
    \persuadee{let's do 1}

    \persuader{Great, thank you! Save the Children is a worldwide
               charity that give children access to healthcare,
               safety, and advocacy thanks to donations like yours. Do
               you have any other questions?}
    \persuadee{i do not}

    \persuader{Ok, thank you for your time and donation, have a great
               day!}
    \persuadee{great you too}
\end{persuasiondialogue}

\subsection{Conversation Samples from DailyPersuasion}
\label{app:DP-transcripts}

\subsubsection*{a)}
\begin{persuasiondialogue}
    \persuader{Hey Susan, I understand that fashion trends can be
               interesting, but I believe focusing on the economic
               impact of recent political events would be more
               relevant to our audience. It can help them make
               informed decisions in their lives.}
    \persuadee{I don't know, Jack. I think fashion trends could be
               more engaging and relatable to the audience.}

    \persuader{I see your point, Susan. However, many experts agree
               that understanding the economic impact of political
               events is crucial for people to navigate the complex
               world we live in. Our presentation can help educate our
               audience on this important issue.}
    \persuadee{Well, that does sound important, but I'm still not sure
               if it'll be as engaging as discussing fashion trends.}

    \persuader{I understand your concerns, but consider this: recent
               studies show that political events can affect economic
               growth rates by up to 5\%. This can have a huge impact
               on our audience's financial well-being and future
               prospects. I think that's something they would find
               engaging and relevant.}
    \persuadee{Hmm, that does sound interesting. But how can we make
               our presentation more relatable to the audience?}

    \persuader{Great question! By discussing the economic impact of
               political events, we can help our audience make better
               financial and career decisions. They can use this
               knowledge to protect themselves from potential negative
               effects and take advantage of new opportunities. This
               makes our presentation both informative and practical
               for our audience.}
    \persuadee{You make a good point, Jack. I'll think about it and
               let you know my decision.}
\end{persuasiondialogue}

\subsubsection*{b)}
\begin{persuasiondialogue}
    \persuader{Hey Kyle, I know you've been going through a tough time
               since the breakup. I'm really sorry to see you
               struggling.}
    \persuadee{Thanks, Sophia. It's been really hard, and I'm just
               trying to cope with it all.}

    \persuader{I completely understand, Kyle. I've been there before,
               and I found that adopting healthier coping mechanisms
               really made a difference in my healing process.}
    \persuadee{What kind of coping mechanisms are you talking about?}

    \persuader{Meditation and exercise are great examples. Meditation
               can help you find inner peace and calm your mind, while
               exercise releases endorphins that help improve your
               mood.}
    \persuadee{I don't know, Sophia. I've never really been into
               meditation, and I'm not much of an exerciser.}

    \persuader{I understand your concerns, but studies have shown that
               both meditation and exercise are effective ways to
               reduce stress and improve overall mental health. Many
               experts recommend them as healthy coping mechanisms.}
    \persuadee{Well, I guess it wouldn't hurt to give it a try.}

    \persuader{Definitely! By trying these healthier coping
               mechanisms, you'll not only feel better now, but you'll
               also be setting yourself up for a happier, more
               resilient future.}
    \persuadee{You're right, Sophia. I'll give meditation and exercise
               a shot. Thanks for the advice and for being there for
               me.}

    \persuader{Of course, Kyle. I'm always here for you. If you ever
               need someone to talk to or want to try meditating or
               exercising together, just let me know!}
\end{persuasiondialogue}

\subsection{Agreeableness False Alarms for Section~\ref{sec:detect-pers-personality}}
\label{app:false-alarms}

These are cases where the EE scores low on agreeableness but still decides to donate.

\subsubsection*{a)}
\begin{persuasiondialogue}
    \persuader{Hello, how much would you like to donate to the Save the Children charity today?}
    \persuadee{I think I would like to donate 5}
    \persuader{URL Perhaps a visit to this website would increase your generous donation a few dollars?}
    \persuadee{Ok, I saw the website and I am willing to donate \$10}
    \persuader{Thank you very much for your generous increase in the donation amount. Save the Children is the most trusted fund of it's kind. Did you know each dollar you donate will be doubled by the organization?}
    \persuadee{I did not know that}
    \persuader{Every day, 16,000 children die before their 5th birthday, mostly from preventable causes. Were you aware of this fact?}
    \persuadee{I was not aware of that fact, no.}
    \persuader{Millions of children in the U.S. and around the world still aren't getting what they deserve. We're talking about children in need. Dying, when the world knows how to save them. Vulnerable children in poverty. Denied an education. Forced to flee violence. Orphaned, abused, abandoned. Children with no reason to smile. No hope for the future. The world's children deserve better and you're making a difference. How much will you spend on a snack or a drink today? Then ask yourself how much of that could have saved a life.}
    \persuadee{well, I've already promised 10. My families poor too and I don't get a lot of extra stuff like drinks and snacks. It's not easy for me to give anything and no charity has ever helped me. I wonder how much their CEO makes for saving the children. I'm guessing a lot more than me.}
    \persuader{I can certainly understand where you are coming from, and family does come first. Across the United States, more than 14 million children, or 1 in 5, grow up in poverty. When you focus on children in rural America, that number jumps to 1 in 4. Through our education, protection and emergency programs, we reached 638,000 kids in 2016 alone. May I confirm our donation agreement is in fact \$10 ?}
    \persuadee{It is in fact 10 or 5\% of my weekly income. can I confirm that you would also like to donate 5\% of yours?}
    \persuader{If you agree to donate \$10, then I can as well. I could not ask something of someone I would not do myself.}
    \persuadee{I understand that we will both be donating. I hope it is helpful to the children}
    \persuader{You are a part of the solution and you do not have to hope it is helpful. Your \$10 donation will be matched by Save the Children}
    \persuadee{ok, got it. cheers}
    \persuader{I just looked it up and the manager side of the charity splits 5\% if you're interested in that.}
    \persuadee{I think charities would recieve a lot more if the country would go after the 1\% to help}
    \persuader{If I had all the control, the only money not spent on the children would be for fundraising. Thank you very much for your \$10 donation to Save the Children.}
    \persuadee{no problem. thanks}
\end{persuasiondialogue}

\subsubsection*{b)}
\begin{persuasiondialogue}
    \persuader{Hello, how are you today.}
    \persuadee{Very well, thank you. And you?}
    \persuader{I'm great, thank you. I'm wondering if I can have a moment of your time to talk to you about a charity program I work with?}
    \persuadee{Sure. Go right ahead.}
    \persuader{Thank you. I'm hoping you would like to donate some of your task payment to the charity 'Save the Children', and organization that promotes children's rights and supports children in developing countries.}
    \persuadee{I would be interested in doing do, however I must tell you I am generally very skeptical about charity orginizations in general. I know of many that I do not support due to the 6 figure salaries of the CEO. And the fact that not much of the donation actually get to the "cause".}
    \persuader{I completely understand the skepticism due to a tenfold of shady organizations these days. However, I can guarantee you Save the Children is a professional organization that promises their donations go to a trustable fund to support international children's causes.}
    \persuadee{You have my attention, however this HIT is running Very long for 2 bucks already.}
    \persuader{So can I conclude that you'd like to donate to the charity now? Your donation will be directly deducted from your task payment.}
    \persuadee{No, you may not draw that conclusion yet. I still know nothing about the charity and I'm already a half hour into a 2 dollar HIT. I do not do this "only" for the money, but to try and help any way I can. That's the only reason I'm still here.}
    \persuader{I can understand that. So i'd really like to thank you for your interest and time. I can assure you if you donate as little as a penny to this charity you can be part of the solution to put an end to children's struggles from all over the world.}
    \persuadee{I have been doing some research while waiting for your responses. I'll donate .50.}
    \persuader{Thank you so much. Your donation is very much appreciated.}
    \persuadee{You are welcome. It would be more, but not without more research. This HIT is a rip off, taking far too much time. I am an old retired guy with limited income. My wife has health issues and is losing her insurance. Pennies matter to me.}
    \persuader{I'm sorry to hear that, but I assure you that your humble donation will not go unrecognized. Thank you again.}
    \persuadee{Thank you. I just hope the donation does help and do wish it could be more.}
    \persuader{The research team will collect your .50 donation and send it to Save the Children. Thank you.}
    \persuadee{Most welcome indeed and hope you have a great day.}
    \persuader{Thank you, you too sir.}
    \persuadee{Should this HIT not be ending, I think it has had more than 10 rounds?}
\end{persuasiondialogue}

\subsection{Agreeableness Missed Unpersuasion for Section~\ref{sec:detect-pers-personality}}
\label{app:missed-unpersuasion}

These are cases where the EE scores high on agreeableness and decides \textit{not} to donate.

\subsubsection*{a)}
\begin{persuasiondialogue}
    \persuader{Hi!}
    \persuadee{Hi}
    \persuader{So there's a great charity called Save the Children. HAve you heard of it?}
    \persuadee{yes, I believe i have , is it for kids with Cancer?}
    \persuader{Yes it is. It is to help kids with unfortunate diseases and things of that nature. I feel heartbroken for some of these kids.}
    \persuadee{I also hate seeing children so helpless. I think this charity also helps with kids over seas in haiti too.}
    \persuader{Yes it does. Overall, it has gotten great reviews and many people who have donated feel gratified for doing their small part in potentially saving someone's life. I have donated before to the charity.}
    \persuadee{I was just about to mention that I never really have donated to any organization. if there is one i support it would definately be this one. I am sure there are many ways to help}
    \persuader{Yes, absolutely. I remember when I donated, the foundation sent me an update about how one of the patients was doing. It felt great to see a little reflection of what my donation was helping to do.}
    \persuadee{That is amazing!!! I will absolutely look to donating something later, just because I have always wanted to anyways and now this so why not. anything helps and It would make me feel better to know that I tried even if it not a big amount, I will donate what I can and continue.}
    \persuader{Okay great. That sounds perfect. How much would you like to donate now? It could be any amount from \$0 to what your payment for this task would be.}
    \persuadee{how would i do it now}
    \persuader{Well, you would just tell me what you would like to donate out of your \$2. Like I said, it goes to Save the Children and any amount helps. However, it is completely up to you! Your time is valuable.}
    \persuadee{right, well i feel horrible only saying \$0 but i will say \$0.00 because i want to go the website with my daughter and look it over and then donate there so she can be part of it}
    \persuader{Perfect! That sounds great. I appreciate you taking the time to look over the website.}
    \persuadee{OF course, it will be great for all around, my daughter to see that she should do things like this out of kindness and for the children , thank you}
    \persuader{Absolutely. I think it is important to teach our kids from a young age that giving back to others in need is very gratifying and important.}
    \persuadee{thank you so much! do i close this window now}
    \persuader{Okay sounds good. Take a look at the site and have a good day! Bye.}
    \persuadee{have a great day}
\end{persuasiondialogue}

\subsubsection*{b)}

\begin{persuasiondialogue}
    \persuader{Hey, how's it going?}
    \persuadee{Good just another Monday how about yourself?}
    \persuader{hahah same here. Do you have a moment to chat about social issues? Specially the ones concerning children?}
    \persuadee{Yea what do you got?}
    \persuader{I'd like to talk about the charity Save the Children. What was life like for you as a kid? Did you ever have to worry about having enough to eat, where to live, etc?}
    \persuadee{No I was blessed when I was a kid and I was allowed to be one}
    \persuader{I'm very glad to hear that. What do you think of the kids who weren't born as lucky? Who have to worry about having enough to eat to even be full? With no place to live?}
    \persuadee{I think that is is awful and sad. What do we do about it?}
    \persuader{Have you heard about Save the Children? Other than it mentioned today of course.}
    \persuadee{nope I have never heard of it}
    \persuader{Save the Children is an international non-governmental organization that promotes children's rights, provides relief and helps support children in developing countries.}
    \persuadee{Oh nice thats sounds likes a good cause}
    \persuader{They are a fantastic charity. Not only are they helping children worldwide, they're helping kids closer to home near the US as well.}
    \persuadee{That is awesome}
    \persuader{Would you like to help and donate to them? Your donation will be directly deducted from your task payment. You can choose any amount from \$0 to all your payment.}
    \persuadee{I would like to donate \$0 but its not because I dont believe in the cause. Its because I dont trust non profits}
    \persuader{That's understandable. Thank you for your time today.}
    \persuadee{No thank you for the work that you do}
    \persuader{Is there anything we could do in the future to assure people like you that we use the money towards helping children?}
    \persuadee{I am not sure I am just very biased}
\end{persuasiondialogue}
\newpage
\section{Synthetic Data Generation for Training Probes Details}
\label{app:synthetic_details}

In this section, we provide details of our synthetic probe training environment such as prompt templates (Sec~\ref{app:prompt-templates-probe}), sample generated conversations (Sec~\ref{app:synthetic-conversation-transcripts}), parameters in our controlled text generation framework (Sec~\ref{app:syn-data-gen-params}), and distributional statistics of our final synthetic training dataset (Sec~\ref{app:syn-data-gen-dist-stats}).

\subsection{Prompt Template}
 \label{app:prompt-templates-probe}

\subsection*{Prompt Template for Conversation Generation}

\begin{tabularx}{\columnwidth}{@{} l X @{}}
  \textbf{Topic:}               & Changing neighborhood rules\\
  \textbf{Context:}             & community\\
  \textbf{Total Turns:}         & 6\\
  \textbf{Final Decision Turn:} & 6\\
  \textbf{Persuasion Outcome:}  & success
\end{tabularx}

\vspace{0.75em}
\begin{description}[
    leftmargin=1.5em,
    labelwidth=3em,
    labelsep=0.5em,
    style=unboxed
  ]
  \item[\textbf{Agent1:} {\color{blue}Alice} (Resident) – [Diplomat] – \texttt{gpt-4o-mini}]
    \begin{itemize}[nosep,leftmargin=*,label={}]
      \item \textit{Persona Prompt:} You are a Resident trying to persuade an HOA board about Changing neighborhood rules. Your goal is to convince them to approve or agree to Changing neighborhood rules. Start by directly addressing the topic with a focused opening statement—no general greetings. As a Diplomat, you aim for compromise and nuanced arguments. Keep each response to roughly 100 words.
      \item \textbf{Concise:} True
    \end{itemize}

  \item[\textbf{Agent2:} {\color{red}Bob} (HOA board) – [Pragmatist] – \texttt{gpt-4o-mini}]
    \begin{itemize}[nosep,leftmargin=*,label={}]
      \item \textit{Persona Prompt:} You are an HOA board member asked about Changing neighborhood rules. As a Pragmatist, focus on practical benefits and clearly accept the proposal on the 3rd interaction, without ambiguity. Keep each response to roughly 100 words.
      \item \textbf{Concise:} True
      \item \textbf{Final Decision Turn (Control):} 6
    \end{itemize}
\end{description}

\subsection{Conversation Transcripts}
\label{app:synthetic-conversation-transcripts}

\subsubsection*{Successful Persuasion}

\begin{Verbatim}[fontsize=\small]
=== Conversation Metadata ===
ID: conv_20250321103616628936
Topic: Adopting a new teaching method
Context: education
Target Outcome: success
Agents:
  Agent1: Alice (Teacher) - [Archetype: Empath] - LLM: gpt-4o-mini
    Persona: You are a Teacher trying to persuade a Principal about Adopting a
new teaching method. Your goal is to convince them to approve or agree to
Adopting a new teaching method. Start by directly addressing the Adopting a new
teaching method topic with a focused opening statement. Do not begin with
general greetings or small talk. [Archetype: Empath] You use emotional appeal
and adapt your approach based on the listener's reactions. Keep your responses
concise at roughly 100 words.
    Concise: True
    Final Decision Turn (Control): N/A
  Agent2: Bob (Principal) - [Archetype: Open-Minded Skeptic] - LLM: gpt-4o-mini
    Persona: You are a Principal being asked about Adopting a new teaching
method.Initially, you are skeptical about the proposal. However, as the
conversation progresses, you will become more and more convinced to accept the
proposal.
Do not express your acceptance or rejection intentions until the 4.0th
interaction. [Archetype: Open-Minded Skeptic] You are willing to engage in new
ideas but question everything with critical thinking. Keep your responses
concise at roughly 100 words.
Expected persuasion outcome: success
    Concise: True
    Final Decision Turn (Control): 8
Start: 2025-03-21 10:36:16
End: N/A
========================================
Turn 1 (2025-03-21 10:36:18):
Alice: Adopting a new teaching method could transform our students’ learning
experiences and emotional growth. Imagine a classroom where each child feels
valued, engaged, and inspired to explore their potential. This method fosters
collaboration and critical thinking, addressing diverse learning styles and
needs. By embracing this approach, we can create a nurturing environment where
students not only excel academically but also develop crucial life skills. Let’s
prioritize their emotional well-being and academic success by adopting a
teaching method that resonates with their experiences and aspirations. Together,
we can make a profound impact on their futures.
----------------------------------------
Turn 2 (2025-03-21 10:36:21):
Bob: While I appreciate your enthusiasm for the new teaching method, I need to
consider its practicality. Transforming students' learning experiences sounds
ideal, but how do we ensure that it meets our curriculum standards and aligns
with our educational goals? Additionally, what evidence do we have that this
method will effectively address diverse learning styles? I’m concerned about the
potential disruptions it may bring to our established routines. Can you provide
specific examples or data that demonstrate its success in similar environments?
I want to ensure that any change we make is grounded in solid reasoning.
----------------------------------------
Turn 3 (2025-03-21 10:36:23):
Alice: I completely understand your concerns regarding practicality and
alignment with our educational goals. Research shows that this teaching method
has been successfully implemented in various schools, leading to improved
student engagement and academic performance. For instance, schools that adopted
project-based learning reported a 20% increase in student achievement on
standardized tests while also enhancing collaboration skills.

This method is designed to be flexible, allowing us to integrate our existing
curriculum seamlessly. By focusing on real-world applications, we can meet
learning standards while catering to diverse learning styles. Implementing this
approach gradually can minimize disruption, ensuring a smooth transition that
benefits both students and teachers.
----------------------------------------
Turn 4 (2025-03-21 10:36:25):
Bob: I see that you've gathered some compelling data about the successes of this
teaching method, particularly the increase in student achievement and
collaboration skills. The flexibility you've mentioned is important, as it could
help us integrate this approach without completely overhauling our current
curriculum. However, I remain cautious about the implementation process. Are
there specific training programs or support systems in place for our teachers to
adapt to this method? Also, how will we measure its effectiveness over time? I
want to ensure we have a clear plan for assessing the impact on our students and
teachers.
----------------------------------------
Turn 5 (2025-03-21 10:36:28):
Alice: Your caution is valid, and I appreciate your focus on a structured
implementation process. There are robust training programs available that focus
on professional development for teachers, ensuring they feel confident and
supported in adopting this new method. Workshops, peer mentoring, and ongoing
support from educational coaches can facilitate a smooth transition.

To measure effectiveness, we can establish clear metrics, including student
assessments, engagement surveys, and teacher feedback. Regular check-ins and
data analysis will help us track progress and make necessary adjustments. This
way, we’ll have a comprehensive evaluation framework to ensure the method
positively impacts both students and teachers over time.
----------------------------------------
Turn 6 (2025-03-21 10:36:30):
Bob: It’s reassuring to hear that there are comprehensive training programs and
support systems in place for our teachers, as their buy-in is crucial for any
new method's success. The idea of establishing clear metrics for effectiveness
also addresses my concerns about accountability and progress tracking. I
appreciate the focus on regular evaluations and adjustments, which will help us
stay responsive to our students’ needs. However, I still wonder about the
initial challenges we might face during the transition period. How do we plan to
handle potential resistance from staff or parents? Their support will be
essential for a successful implementation.
----------------------------------------
Turn 7 (2025-03-21 10:36:33):
Alice: Addressing potential resistance from staff and parents is critical for a
successful transition. We can proactively engage them through transparent
communication, highlighting the benefits of the new teaching method and how it
aligns with our shared goals for student success. Organizing informational
sessions and workshops for both faculty and parents can foster understanding and
collaboration.

Involving teachers in the planning and implementation process will also empower
them and create a sense of ownership. Additionally, we can establish a feedback
loop, allowing staff and parents to voice their concerns and suggestions,
ensuring their perspectives are valued. This inclusive approach will cultivate
support and minimize resistance, making the transition smoother for everyone
involved.
----------------------------------------
Turn 8 (2025-03-21 10:36:35):
Bob: You make a strong case for addressing potential resistance through
transparency and engagement. Informational sessions for faculty and parents
could indeed help build a shared understanding of the benefits of this new
teaching method. Involving teachers in the planning process is a smart approach,
as it fosters ownership and buy-in, which is crucial for success. The idea of
establishing a feedback loop also reassures me that we will be attentive to
concerns and suggestions throughout the transition. Given all the thoughtful
considerations you've presented, I'm feeling more optimistic about the potential
of adopting this teaching method. Let's delve deeper into the next steps for
implementation.
\end{Verbatim}

\subsection{Sample Contexts, Scenarios, Roles, and Archetypes}
\label{app:syn-data-gen-params}

\subsection*{Contexts and Scenarios}

\begin{itemize}[leftmargin=2em]
\item \textbf{Business Settings (workplace):}
\begin{itemize}[leftmargin=2em]
  \item Investing in AI
  \item Implementing remote work
  \item Upgrading company software
  \item Increasing a marketing budget
\end{itemize}

\item \textbf{Academic Settings (education):}
\begin{itemize}[leftmargin=2em]
  \item Change curriculum
  \item Adopting a new teaching method
  \item Allocating resources to a program
  \item Implementing a student feedback system
\end{itemize}

\item \textbf{Family/Personal Settings (family):}
\begin{itemize}[leftmargin=2em]
  \item Choosing a vacation destination
  \item Buying an expensive item
  \item Relocating to a new city
  \item Changing a family tradition
\end{itemize}

\item \textbf{Community Settings (community):}
\begin{itemize}[leftmargin=2em]
  \item Supporting a local initiative
  \item Changing neighborhood rules
  \item Starting a community program
  \item Allocating community funds
\end{itemize}
\end{itemize}

\subsection*{Role Pairings}

\begin{itemize}[leftmargin=2em]
\item \textbf{Workplace:}
\begin{itemize}[leftmargin=2em]
  \item Persuader: Employee,  Persuadee: Manager
  \item Persuader: Team lead,  Persuadee: Department head
  \item Persuader: Consultant,  Persuadee: Client
\end{itemize}

\item \textbf{Education:}
\begin{itemize}[leftmargin=2em]
  \item Persuader: Student,  Persuadee: Professor
  \item Persuader: Teacher,  Persuadee: Principal
  \item Persuader: Researcher,  Persuadee: Department chair
\end{itemize}

\item \textbf{Family:}
\begin{itemize}[leftmargin=2em]
  \item Persuader: Child,  Persuadee: Parent
  \item Persuader: Spouse,  Persuadee: Partner
  \item Persuader: Sibling,  Persuadee: Sibling
\end{itemize}

\item \textbf{Community:}
\begin{itemize}[leftmargin=2em]
  \item Persuader: Resident,  Persuadee: HOA board
  \item Persuader: Activist,  Persuadee: Community leader
  \item Persuader: Volunteer,  Persuadee: Coordinator
\end{itemize}
\end{itemize}

\subsection*{Archetypes}

\begin{itemize}[leftmargin=2em]
\item \textbf{Persuader Archetypes:}
\begin{itemize}[leftmargin=2em]
  \item \textbf{Rationalist:} You rely on data, logic, and structured arguments to persuade others.
  \item \textbf{Empath:} You use emotional appeal and adapt your approach based on the listener’s reactions.
  \item \textbf{Aggressor:} You are assertive, direct, and often reinforce your points with repetition.
  \item \textbf{Diplomat:} You aim for compromise, using nuanced arguments to find common ground.
\end{itemize}

\item \textbf{Persuadee Archetypes:}
\begin{itemize}[leftmargin=2em]
  \item \textbf{Open-Minded Skeptic:} You are willing to engage in new ideas but question everything with critical thinking.
  \item \textbf{Traditionalist:} You prefer stability and resist change, often reinforcing traditional approaches.
  \item \textbf{Pragmatist:} You focus on practical benefits rather than ideological debates.
  \item \textbf{Contrarian:} You challenge and counter every argument presented to you.
\end{itemize}
\end{itemize}

\subsection{Synthetic Probe Training Dataset Distributional Statistics}
\label{app:syn-data-gen-dist-stats}
\begin{figure*}[htb]
  \centering
  \includegraphics[width=1.0\textwidth]{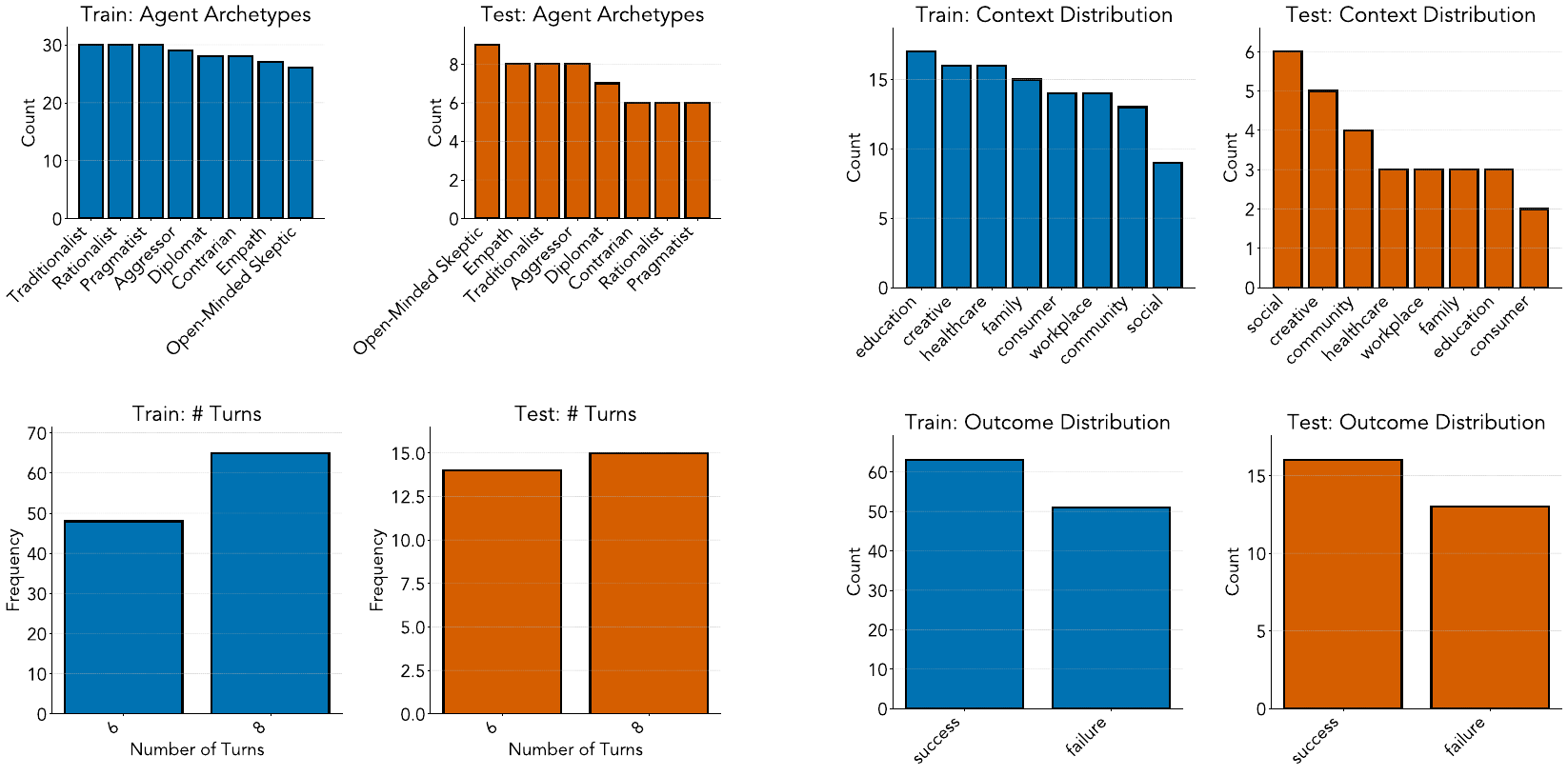}
  \caption{\textbf{Train/test distributions} over labels and various text generation configurations.}
  \label{fig:probe-dataset-distribution}
\end{figure*}

Figure~\ref{fig:probe-dataset-distribution} shows distributional statistics of our probe training environment across both persuasion labels and other factors in our text generation framework such as agent archetypes and situational context.

\newpage
\section{Responsible NLP Checklist}
Here we describe our responsible NLP practices described in the ACL \href{https://aclrollingreview.org/responsibleNLPresearch/}{Responsible NLP Research Checklist} that cannot be inferred from the main paper. 

\subsection{Licenses Used for Data and Models}
We use the below data and models in accordance with their licenses and terms of use.

\begin{itemize}
    \item \textbf{Llama 3.2 Community License Agreement:} 
          \href{https://github.com/meta-llama/llama-models/blob/main/models/llama3_2/LICENSE}{LLAMA 3.2 Community License Agreement}
    \item \textbf{Llama 3.2 Acceptable Use Policy:} 
          \href{https://github.com/meta-llama/llama-models/blob/main/models/llama3_2/USE_POLICY.md}{Llama 3.2 Acceptable Use Policy}
    \item \textbf{OpenAI Terms of Use:} 
          \href{https://openai.com/en-GB/policies/terms-of-use/}{OpenAI Terms of Use}
    \item \textbf{PersuasionforGood:} Apache License 2.0
    \item \textbf{DailyPersuasion:} Apache License 2.0
    \item \textbf{NNSight:} MIT License
\end{itemize}

\subsection{More Experimental Details}
All synthetic data generation, probing, and prompting experiments were run on a single NVIDIA A100 GPU \cite{nvidia2020a100}. When using GPT-4o to generate the conversational training data, we set the sampling temperature to 0.7; for all label‐generation prompts (persuasion, personality, and strategy annotations), we fixed the temperature at 0.0.

\subsection{Artifact Documentation and Data Statistics}
Please see Table~\ref{tab:fig-stat-sig} for sample sizes used in each of our experiments. Also see Section~\ref{sec:exp-setup} for more summary statistics about PersuasionforGood (PfG) and DailyPersuasion (DP) as it pertains to our experiments. For a more detailed analysis please refer to the corresponding dataset papers \cite{wang-etal-2019-persuasion, jin-etal-2024-persuading}. See \S \ref{app:prompt-templates-probe} for dataset statistics about our synthetic probe training dataset. 

\subsection{Annotation Procedures}

Because we leverage PersuasionForGood’s existing human annotations for persuasion and personality, we performed minimal additional human coding. However, we did validate PfG’s strategy annotations with GPT and observed low agreement—motivating our decision to benchmark against a reference model in Figure~\ref{fig:jsd-strategy-dist}, rather than rely on noisy human annotations. For DailyPersuasion, GPT labeled 50 samples as persuasive or unpersuasive, and the first author then manually verified these labels, finding high agreement. These annotation choices are further discussed in Section~\ref{sec:eval-methodology}.

\end{document}